\documentclass{article} 
\usepackage{iclr2026_conference,times}

\usepackage{amsmath,amsfonts,bm}

\def\eqref#1{equation~\ref{#1}}

\def\1{\bm{1}}

\DeclareMathAlphabet{\mathsfit}{\encodingdefault}{\sfdefault}{m}{sl}
\SetMathAlphabet{\mathsfit}{bold}{\encodingdefault}{\sfdefault}{bx}{n}

\usepackage{hyperref}
\usepackage{url}

\usepackage[utf8]{inputenc} 
\usepackage[T1]{fontenc}    
\usepackage{hyperref}       
\usepackage{url}            
\usepackage{booktabs}       
\usepackage{amsfonts}       
\usepackage{nicefrac}       
\usepackage{microtype}      
\usepackage{xcolor}         

\usepackage{makecell}
\usepackage{array}

\RequirePackage{fancyhdr}
\RequirePackage{algorithm}
\RequirePackage{algorithmic}
\RequirePackage{eso-pic} 
\RequirePackage{forloop}
\usepackage{amssymb}
\usepackage{mathtools}
\usepackage{amsthm}
\usepackage{cleveref}
\crefname{equation}{Eq.}{Eqs.}
\crefname{table}{Table}{Tables}
\crefname{figure}{Figure}{Figures}
\crefname{section}{Section}{Sections}
\crefname{algorithm}{Algorithm}{Algorithms}
\theoremstyle{plain}

\theoremstyle{definition}

\theoremstyle{remark}

\usepackage{colortbl}
\usepackage{adjustbox}
\usepackage{url}
\usepackage{multirow,multicol,xspace}
\usepackage{float}
\usepackage{graphics}
\usepackage{paralist}
\usepackage{wrapfig}
\usepackage[normalem]{ulem}
\usepackage{pgfplots}
\pgfplotsset{width=8cm,compat=1.17} 
\usepackage{pifont}
\usepackage{subcaption,ragged2e}
\usepackage{caption}
\usepackage{graphicx}
\usepackage{wasysym}   

\usepackage{minted} 
\setminted{
  breaklines,
  breakanywhere,
  autogobble,
  linenos,
  numbersep=6pt,
  fontsize=\footnotesize,
  tabsize=2
}

\newcommand{\method}{{DeepEvolve}\xspace}
\newcommand{\mt}{{Molecular Translation}\xspace}
\newcommand{\mol}{{Molecular Prediction}\xspace}
\newcommand{\cp}{{Circle Packing}\xspace}
\newcommand{\be}{{Burgers' Equation}\xspace}
\newcommand{\pd}{{Parkinson's Disease}\xspace}
\newcommand{\nuclei}{{Nuclei Image}\xspace}
\newcommand{\ov}{{Open Vaccine}\xspace}
\newcommand{\pp}{{Polymer Prediction}\xspace}
\newcommand{\usp}{{USP P2P}\xspace}

\usepackage{listings}

\definecolor{codebg}{rgb}{0.95,0.95,0.95}
\definecolor{commentgray}{rgb}{0.4,0.4,0.4}
\definecolor{stringgreen}{rgb}{0.2,0.6,0.2}

\lstdefinestyle{compactpython}{
  language=Python,
  basicstyle=\fontsize{6.5pt}{7.5pt}\selectfont\ttfamily,
  backgroundcolor=\color{codebg},
  keywordstyle=\color{blue},
  commentstyle=\color{commentgray}\itshape,
  stringstyle=\color{stringgreen},
  showstringspaces=false,
  breaklines=true,
  frame=single,
  tabsize=2,
  captionpos=b,
  aboveskip=3pt,
  belowskip=3pt
}

\title{Scientific Algorithm Discovery by Augmenting AlphaEvolve with Deep Research}

\author{
Gang Liu$^{1}$, \quad Yihan Zhu$^{1}$ \quad Jie Chen$^2$ \quad  Meng Jiang$^1$ \\
$^1$University of Notre Dame \quad $^2$ MIT-IBM Watson AI Lab, IBM Research \\
\texttt{\{gliu7, yzhu25, mjiang2\}@nd.edu}, \quad
\texttt{chenjie@us.ibm.com}
}

\newcommand{\algsym}{f}

\iclrfinalcopy 
\begin{document}

\maketitle

\begin{abstract}
Large language models hold promise as scientific assistants, yet existing agents either rely solely on algorithm evolution or on deep research in isolation, both of which face critical limitations. Pure algorithm evolution, as in AlphaEvolve, depends only on the internal knowledge of LLMs and quickly plateaus in complex domains, while pure deep research proposes ideas without validation, resulting in unrealistic or unimplementable solutions. We present \method, an agent that integrates deep research with algorithm evolution, uniting external knowledge retrieval, cross-file code editing, and systematic debugging under a feedback-driven iterative loop. Each iteration not only proposes new hypotheses but also refines, implements, and tests them, avoiding both shallow improvements and unproductive over-refinements. Across nine benchmarks in chemistry, mathematics, biology, materials, and patents, \method consistently improves the initial algorithm, producing executable new algorithms with sustained gains. By bridging the gap between unguided evolution and research without grounding, \method provides a reliable framework for advancing scientific algorithm discovery. Our code is available at \url{https://github.com/liugangcode/deepevolve}.
\end{abstract}

\label{sec:introduction}
\section{Introduction}
Large language models (LLMs) are emerging as foundation models for building AI scientists, automating processes such as lab work, mathematical discovery, and ML research~\citep{boiko2023autonomous,chan2024mle}. Many scientific problems are difficult to solve but easy to evaluate~\citep{romera2024mathematical}, raising hope that LLMs can drive algorithm discovery through reasoning, planning, and execution. Recent progress shows advances in ML benchmarks~\citep{chan2024mle}, mathematical discovery~\citep{novikov2025alphaevolve}, and experimental design~\citep{boiko2023autonomous}. However, it is still challenging for LLM-based agents to push algorithmic frontiers by not only generating new hypotheses~\citep{gottweis2025towards} but also implementing them as working code.

\begin{figure*}[t]
    \centering
    \includegraphics[width=0.9\textwidth]{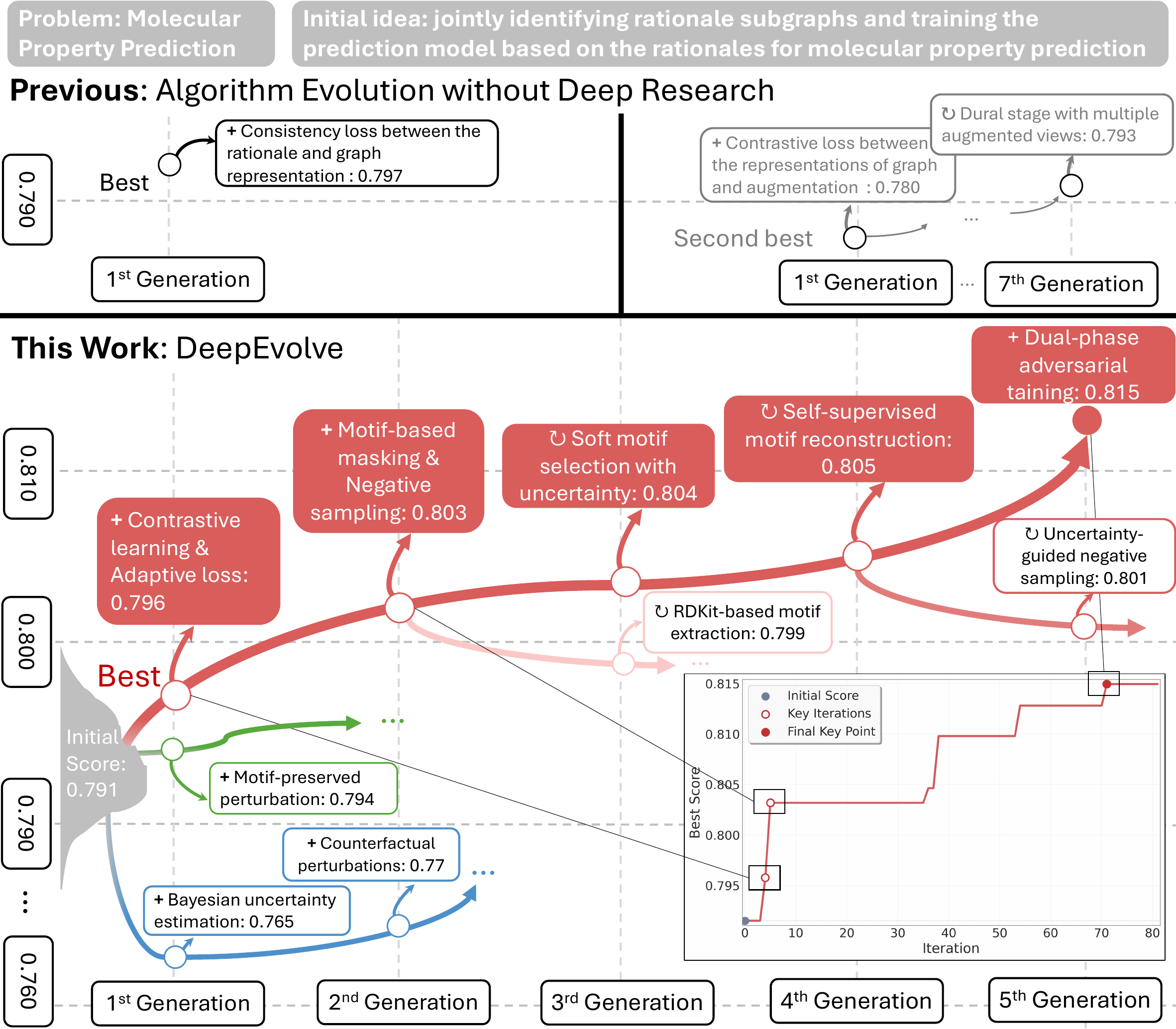}
    \caption{The top panel shows AlphaEvolve-style pure algorithm evolution without deep research, where the best improvement appears in the first generation and later iterations have marginal gains. The bottom panel shows \method, which integrates deep research. \method avoids shallow or excessively deep but unproductive evolutions, achieving sustained progress with clear performance jumps at key iterations. $+$ denotes adding a new idea, and $\circlearrowright$ denotes refining a previous idea.
    }
    \label{fig:idea-evo}
\end{figure*}

The combination of hypothesis generation with code execution and evaluation has been explored in systems such as FunSearch~\citep{romera2024mathematical} and AlphaEvolve~\citep{novikov2025alphaevolve}, with the latter achieving breakthroughs in $4 \times 4$ matrix multiplication. AlphaEvolve uses an ensemble of LLMs to generate code that encodes new scientific hypotheses. However, its generalization to broader domains such as chemistry, biology, and materials remains uncertain.
These domains present vast, unbounded search spaces, where relying solely on LLMs themselves is unlikely to yield substantive algorithmic advances.
A preliminary study of molecular property prediction is shown at the top of~\cref{fig:idea-evo}. Pure algorithm evolution with AlphaEvolve\footnote{The code of AlphaEvolve is unavailable; we follow an open-source reproduction~\citep{openevolve}.} yields limited improvement ($0.791 \rightarrow 0.797$), only 0.6\% after 100 iterations. Surprisingly, the best algorithm appears in the first generation evolved from the initial algorithm, outperforming the other 24 candidates with deeper generations. Some deeply evolved algorithms, including the second-best one, show only marginal improvements after multiple refinements of the initial idea. 

From the figure, we find that high-quality idea generation can be a bottleneck for algorithm evolution in broader scientific domains. To address this, we augment the evolution system with deep research, a framework designed for intensive knowledge work that requires thorough and reliable retrieval from the internet. General deep research methods~\citep{xu2025comprehensive} synthesize information from diverse online sources for scientific hypothesis generation but lack feedback from hypothesis testing. This may lead to proposals that are too difficult or unrealistic to implement. To address this limitation, we perform deep research on a specific algorithm, accompanied by inspiring algorithms that have been successfully implemented in past discoveries. We instruct deep research to generate research proposals with pseudo-code that are easy to implement in the early stages, while moving toward higher-impact ideas in later stages. 
Proposals for an algorithm often involve modifying multiple code files, such as those for data preprocessing or model architecture. This requires the coding agent to parse and analyze across files, a capability added to our design but absent in AlphaEvolve, which substantially increases coding difficulty.
A debugging agent is thus introduced to resolve errors during execution, further improving the success rate of algorithmic implementation (\cref{tab:debug}). Finally, the evaluation function tests the algorithm proposal and provides feedback to deep research for the next proposal. As shown at the bottom of~\cref{fig:idea-evo}, this approach produces clear improvements over both the initial algorithm and pure algorithmic evolution. Unlike shallow evolutions or overly deep but marginal ones, deep research balances depth and yields clear performance jumps at key iterations.

In this work, we propose \method to orchestrate algorithmic deep research, implementation, evaluation, and evolution. The workflow, shown in~\cref{fig:overview-method}, has six components. The first three generate a research proposal by planning research questions, searching for answers online, and composing a proposal. This is then used as a prompt for the coding agent, which performs cross-file edits and multiple rounds of debugging. Each algorithm is evaluated and stored in a database that serves as long-term memory, providing candidates and inspiration for the next round of evolution.

We benchmark nine scientific problems across chemistry, mathematics, biology, materials, and patent domains, covering diverse data modalities such as molecules, geometries, partial differential equations, and images (\cref{tab:benchmarks}). Results show consistent improvements over existing algorithms, generating original and promising new methods (\cref{fig:idea_evaluate_with_llm}) with high performance scores (\cref{tab:algo_indicator}).

\section{Problem Definition for Algorithm Discovery}
Let $P = (D, g)$ denote a scientific problem in domains such as mathematics, chemistry, or biology.  
Each problem has evaluation data $D = \{(q_i, a_i)\}_{i=1}^N$, where $q_i$ are questions and $a_i$ are ground-truth answers, and an evaluation function $g$ that compares the ground-truth answers with predicted answers.  
The score is computed as $ s = g(\{a_i\}_{i=1}^N, \{\hat{a}_i\}_{i=1}^N). $
Here $\hat{a}_i$ are the outputs of an algorithm $\algsym: Q \to A$ that maps each question $q_i$ to an answer $\hat{a}_i = \algsym(q_i)$.  
Both computation and evaluation should be completed within bounded time (e.g., minutes or hours). 
We define a textualization function $\tau$ that converts structured objects into text.  
For example, $\tau(P)$ is the problem description as $\tau_P$ and $\tau(\algsym)$ is the algorithm description as $\tau_h$. The goal of algorithm discovery is to optimize $\algsym$ for higher $s$.

A problem instance in mathematics and geometry is the circle packing. The evaluation is to maximize the sum of radii for $n$ circles placed within a unit square. This can be formalized as a constrained problem $P$. The algorithm $\algsym$ is a Sequential Least Squares Programming (SLSQP) solver, as shown in an open-source reproduction of AlphaEvolve~\citep{novikov2025alphaevolve,openevolve}. Different evaluation data correspond to different values of \( n \), such as \( n = 26, 27, \ldots \).

A second example is molecular property prediction. The goal is to develop ML algorithms that train models to generalize well. They should also yield interpretable predictions for each molecule. We study automated discovery of such algorithms across domains using research and coding agents.

\section{DeepEvolve for Algorithm Discovery}
\begin{figure*}[t]
    \centering
    \includegraphics[width=0.9\textwidth]{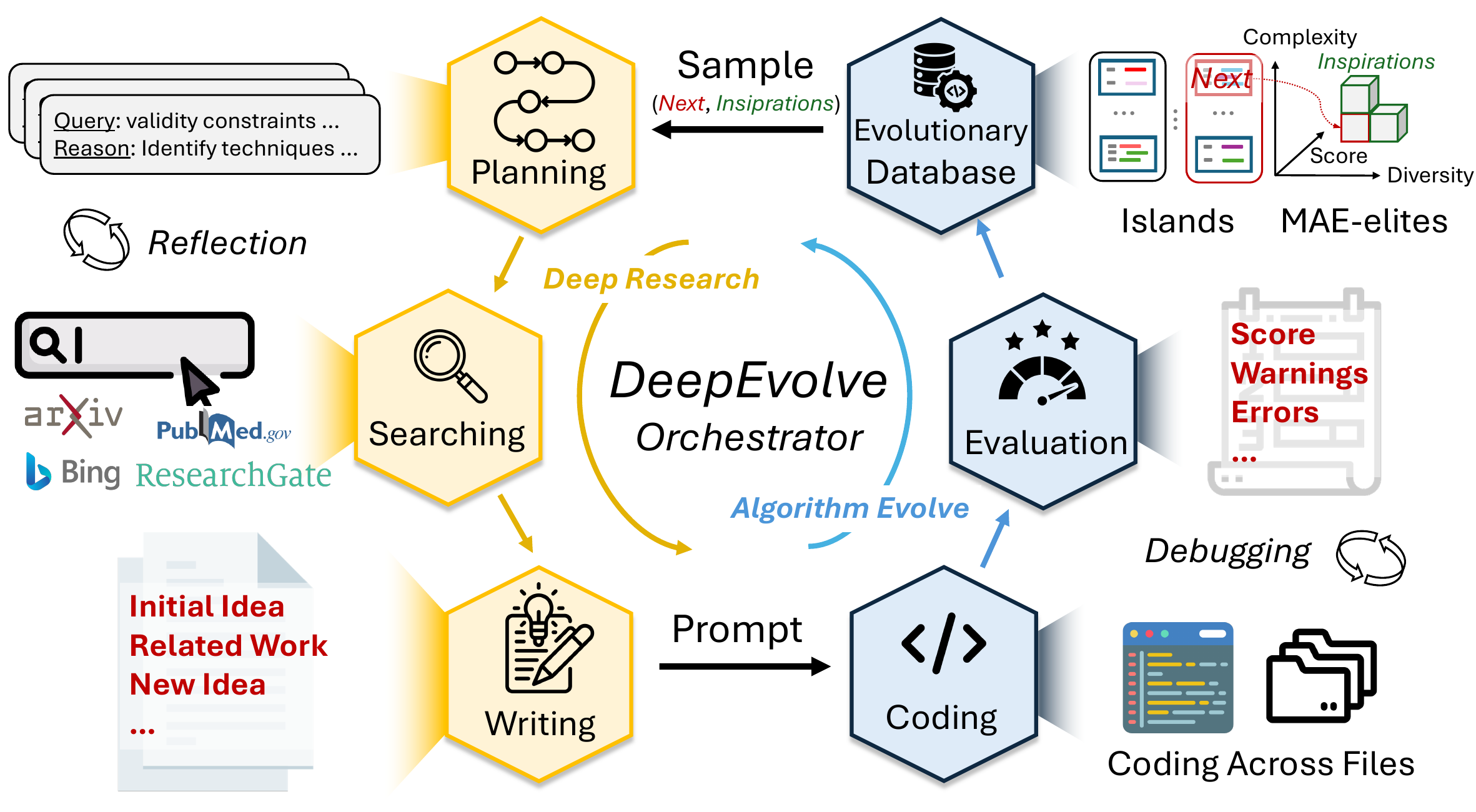}
    \caption{\method is structured around six collaborative modules that alternate between deep research and algorithm evolution. Deep research generates informed hypotheses through planning, retrieval, and synthesis, while algorithm evolution translates these hypotheses into code, evaluates them, and applies evolutionary strategies for selection. }
    \label{fig:overview-method}
\end{figure*}

\method takes as input three things: a problem $P$, an initial algorithm $\algsym$, and user instructions $u$. From these, \method produces an updated algorithm. For a fixed problem and user instruction, we can think of an update operator that takes the current algorithm and returns a new one. This operator is built from six modules, applied in sequence: plan, search, write, code, evaluation, and evolutionary selection. Together, they transform the algorithm in a systematic way. The algorithm evolves by repeatedly applying this update operator. Starting with the initial version $\algsym^{(0)}=\algsym$, each new version is produced from the previous one. After $K$ rounds, we obtain a final candidate $\algsym^{(K)}$. The best algorithm is chosen from all the intermediate versions $\{\algsym^{(0)}, \algsym^{(1)}, \dots, \algsym^{(K)} \}$ by selecting the one that achieves the highest evaluation score on the given problem. 
In the following subsection, we first describe how the input context is built~\cref{subsec:method-init}. We then introduce each component in~\cref{subsec:method-design} corresponding to~\cref{fig:overview-method}, detailing the synergy between deep research and algorithm evolution.

\subsection{Input of Problem, Algorithms, and Instructions}\label{subsec:method-init}

\textbf{Problem as Input.} The input context of problem $ P = (g, \mathcal{D}, \tau_P)$ includes three parts: the evaluation function $g$ implemented as code, the evaluation data $ \mathcal{D}$, and a textual problem description $\tau_P$. Evaluation metrics associated with $g$ are summarized in~\cref{tab:benchmarks}. Given $g$ and $\mathcal{D}$, the optimization direction of the algorithm can be specified. The problem description $\tau_P$ consists of one or more paragraphs that define the task, relevant terminology, notations, equations, and evaluation metrics. 

\textbf{User Instructions.} The user instructions $u$ contain a textual specification of user-defined requirements, providing additional guidance for algorithm evolution. While the evaluation metrics $g$ and data $\mathcal{D}$ determine the primary optimization objective, users may express auxiliary preferences or constraints such as desired research directions (e.g., efficiency, interpretability, generalizability), available software dependencies, hardware constraints, and runtime budgets.

\textbf{Algorithm as Input.} 
The algorithm $\algsym$ consists of both the code implementation and its textual description $\tau_h$. Compared to AlphaEvolve~\citep{novikov2025alphaevolve}, we consider the algorithm implementation spanning multiple files with an entry point that computes the outputs for evaluation. 
Each algorithm description $\tau_h$ includes the motivation, a summary, pseudo-code, the performance $s$, and qualitative assessments such as originality, future potential, and implementation difficulty. 

\subsection{Framework Designs}\label{subsec:method-design}

In an iteration from $t$ to $t+1$, we start from a candidate algorithm $\algsym$ together with a set of inspiring algorithms $\{\algsym^\mathrm{insp}_1, \algsym^\mathrm{insp}_2, \dots, \algsym^\mathrm{insp}_n\}$ and their evaluations to conduct deep research. This differs from a direct implementation~\citep{xu2025comprehensive}, which brainstorms ideas without feedback. After proposing a new algorithm, it is implemented with functions distributed across multiple files and supported by automatic debugging. In contrast, AlphaEvolve~\citep{novikov2025alphaevolve} designs algorithms directly with LLMs, evolves code within a single file, and lacks a code correction mechanism.

\textbf{Algorithmic Deep Research.}
The planning step generates a small set of research questions that guide the direction of the next improvement.
The agent is instructed to be more exploratory if the algorithm has already undergone multiple updates. 
These questions are then searched on websites, including sources such as PubMed and arXiv, and the results are summarized in a few paragraphs.
Finally, a writing agent proposes a new algorithm by integrating the retrieved evidence with the input context (i.e., problem, algorithm, and inspirations). It is instructed to compare different methods and identify promising directions. 
A group of new ideas is generated with self-evaluation, and the most promising one is chosen as the final proposal based on the current evolutionary progress. In early stages, it prioritizes feasible ideas, while in later generations it emphasizes higher-impact ideas. Finally, it writes a short proposal for the new algorithm, including pseudo-code to guide the implementation.

\textbf{Algorithmic Implementation.}
We use a coding agent to implement the proposed algorithm.
It parses multi-file codebases using delimiters. It then localizes the minimal set of code regions that require modification and applies targeted updates to implement the proposed algorithm. 
However, it is easy for new code to contain bugs, especially when modifying different files such as those for data preprocessing and model architecture. During execution, error and warning messages provide valuable information for debugging. Therefore, we introduce a debugging agent to handle failures based on program execution feedback. Given a budget (e.g., five attempts), if execution remains unsuccessful after debugging, the algorithm is assigned a score of zero.

\textbf{Evaluation and Evolutionary Database.}
The algorithm is scored ($s>0$) once it is successfully executed and evaluated. We add it with the score to a database, which is maintained with evolutionary methods for sampling the next candidate and inspiring algorithms.
We use island-based populations~\citep{tanese1989distributed} as the candidate pool for the next iteration. At each step, we sample an island and then select $\algsym$ from it, favoring high-score candidates while retaining exploration. For inspirations, MAP-Elites~\citep{mouret2015illuminating} samples nearby algorithms of $\algsym$ based on three features: performance score, code diversity, and code complexity. These features are mapped to cells in a grid, and neighboring cells are used as inspiration for future candidates $\algsym$.

\paragraph{Reflection}
The reflection mechanism is applied in both algorithmic deep research and implementation as a quick checkpoint for potential issues. For deep research, a reflection agent decides whether to continue planning, continue searching, or update the writing report, subject to a maximum number of reflections. For coding, the agent performs self-reflection to check whether its code aligns with the proposed algorithm and to detect potential syntax errors.

In \method, algorithmic deep research, implementation, and evaluation are coupled across multiple iterations. Deep research alone provides knowledge but no tested progress, while implementation and iteration alone explore ideas blindly without grounding in recent research. By linking the two, the process mirrors human discovery: informed by existing knowledge, tested through implementation, refined with feedback, and improved through repeated cycles. To integrate the iterations more compactly, we instruct the deep research agents based on evolutionary progress (early or mature) and algorithmic history with evaluation feedback. We also use multiple checkpoints (e.g., code modification, self-reflection, debugging) for the coding agent to verify whether its implementation aligns with the proposed algorithm. Empirically, we study how deep research, implementation, and evaluation reinforce each other through evolutionary optimization in~\cref{subsec:exp-iterate-synergy}.

\section{Experiments}\label{sec:experiment}
We investigate three research questions (RQs): RQ1: Can \method discover new algorithms that improve both effectiveness and efficiency across diverse tasks? RQ2: How do the deep research and coding agents interact during the discovery process? RQ3: We conduct ablations and case studies to examine the designs and performance of \method.

\begin{table*}[t]
\centering
\caption{Benchmark tasks, data types, domains, and evaluation metrics. New scores are used for evaluation such that higher values indicate better performance.}
\label{tab:benchmarks}
\resizebox{\textwidth}{!}{
\begin{tabular}{>{\raggedright\arraybackslash}p{1.9cm}
                >{\raggedright\arraybackslash}p{3.3cm}
                >{\raggedright\arraybackslash}p{2.2cm}
                >{\raggedright\arraybackslash}p{2.0cm}
                >{\raggedright\arraybackslash}p{3.2cm}
                >{\raggedright\arraybackslash}p{3.2cm}
                >{\raggedright\arraybackslash}p{2.0cm}}
\toprule
\textbf{Problem} & \textbf{Description} & \textbf{Data Type} & \textbf{Domain} & \textbf{Original Metric} & \textbf{New Score} & \textbf{Source} \\
\midrule
\mol & Molecular property prediction & Small molecule & Chemistry & AUC over multiple model initializations & $0.5 \cdot \mathrm{AUC}_{\mathrm{mean}} + 0.5 \cdot \mathrm{AUC}_{\mathrm{std}}$ & OGB~\citep{hu2020open} \\
\addlinespace[0.3em]
\mt & Image-to-text translation of chemical structures & Image--molecule pair & Chemistry & Levenshtein distance & $1 - \text{Levenshtein distance}$ & Kaggle~\citep{bms2021} \\
\addlinespace[0.3em]
\cp & Packing circles inside a unit square to maximize sum of radii & Geometry & Mathematics & Mean sum of radii with 26 to 32 circles & Same as Original & AlphaEvolve \& Erich's Packing Center~\citep{novikov2025alphaevolve} \\
\addlinespace[0.3em]
\be & Solving Burgers' equation & Partial Differential Equation & Mathematics & Normalized RMSE (nRMSE) & $\frac{1}{\mathrm{nRMSE} \cdot 10^3}$ & CodePDE~\citep{li2025codepde} \\
\addlinespace[0.3em]
\pd & Disease progression prediction & Time series & Biology & Symmetric Mean Absolute Percentage Error (SMAPE) & Same as Original & Kaggle~\citep{kirsch2023amp}\\
\addlinespace[0.3em]
\nuclei & Nuclei segmentation from images & Image & Biology & Mean average precision (mAP) & Same as Original & Kaggle~\citep{goodman2018dsb} \\
\addlinespace[0.3em]
\ov & mRNA vaccine degradation prediction & mRNA sequence & Biology & Mean column-wise RMSE (MCRMSE) & $\frac{1}{1 + \mathrm{MCRMSE}}$ & Kaggle~\citep{das2020openvaccine} \\
\addlinespace[0.3em]
\pp & Prediction of polymer properties & Polymer & Materials & Weighted MAE (wMAE) and $R^2$ & $\frac{1}{1 + \mathrm{wMAE}} \cdot 0.5 + R^2 \cdot 0.5$ & Kaggle~\citep{liu2025neurips} \\
\addlinespace[0.3em]
\usp & Phrase-level semantic matching in patents & Text & Patent & Pearson correlation & Same as Original & Kaggle~\citep{cenkci2022uspppm} \\
\bottomrule
\end{tabular}
}
\end{table*}

\begin{figure*}[t]
    \centering
    \includegraphics[width=0.95\textwidth]{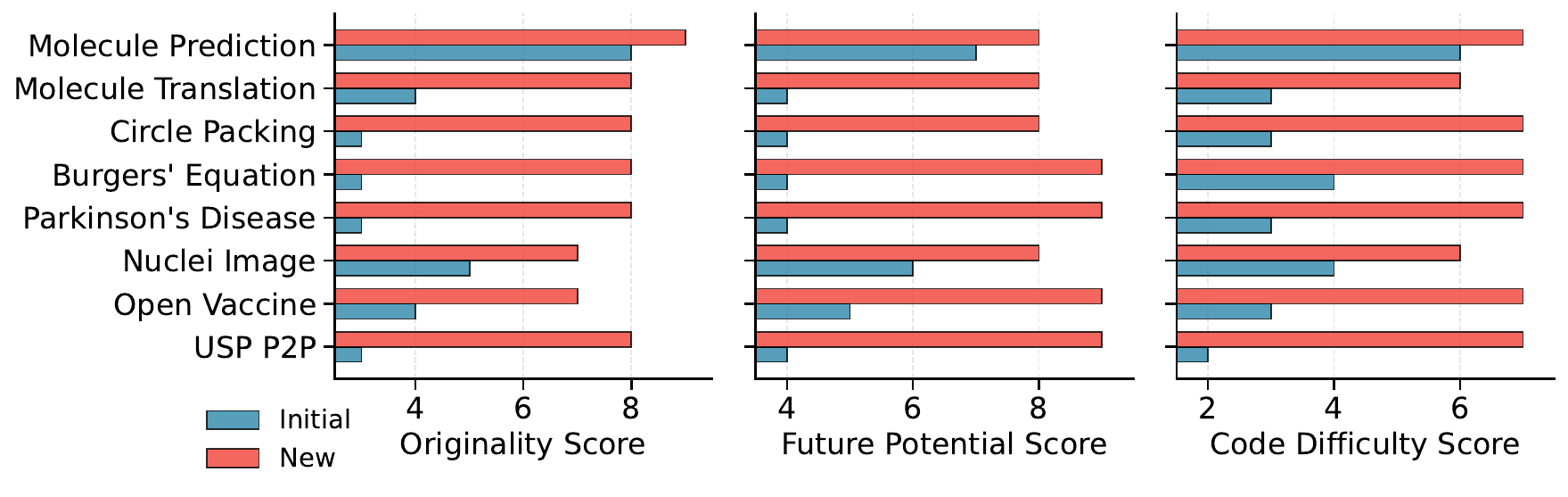}
    \caption{Evaluation of the idea from initial and new algorithms with LLM-as-a-judge.}
    \label{fig:idea_evaluate_with_llm}
\end{figure*}

\subsection{Set-ups}
We include nine research problems spanning chemistry, mathematics, biology, and materials as summarized in~\cref{tab:benchmarks}. These problems involve diverse data modalities, including molecules, images, mRNA, text, time series, geometric structures, and multi-modal inputs. For consistent evaluation, we standardize evaluation metrics (e.g., AUC-ROC, RMSE, precision, Pearson correlation) defined in each problem into a common form as the new scores, where higher values indicate better performance.

For each problem, we designate an initial algorithm as the baseline and apply \method to optimize and generate new algorithms. For the molecule and polymer tasks, we improve the graph rationalization method GREA~\citep{liu2022graph} in different directions specific to each problem. For the circle packing problem, we adapt the SLSQP algorithm from OpenEvolve~\citep{openevolve}, an open-source implementation of AlphaEvolve. For the Burgers equation, we use the baseline provided by CodePDE~\citep{li2025codepde}. For problems derived from Kaggle competitions, including molecular translation, Parkinson’s disease progression, nuclei image segmentation, Open Vaccine, and USP P2P, we use baseline solutions provided by competition participants. More details are in~\cref{sec:add-benchmark}. 

To discover new algorithms, we define the primary optimization objective as the new scores in~\cref{tab:benchmarks}, with efficiency specified as a secondary objective in the prompt. The algorithm development process is constrained to a 30-minute time budget and a single GPU (2080-Ti or A6k). We evaluate both baseline and generated algorithms using quantitative metrics and qualitative analysis.

\begin{table}[t]
\centering
\caption{Quantitative comparison of new algorithms discovered by \method with the initial ones in terms of effectiveness (new scores; see~\cref{tab:benchmarks}) and efficiency (runtime in minutes). Efficiency is not the primary optimization objective in \method; it could be included in the user query.}
\label{tab:algo_indicator}
\resizebox{0.96\textwidth}{!}{
\begin{tabular}{l c c c c c c}
\toprule
Problem 
& \multicolumn{3}{c}{Performance with New Scores ($\uparrow$)} 
& \multicolumn{3}{c}{Runtime in Minutes} \\
\cmidrule(lr){2-4} \cmidrule(lr){5-7}
& \makecell{Initial\\Algorithm}
& \makecell{New\\Algorithm}
& \makecell{Improvement\\(\%)} 
& \makecell{Initial\\Algorithm}
& \makecell{New\\Algorithm}
& \makecell{Reduced Time \\(Minutes)} \\
\midrule
\mol              & 0.7915 & 0.8149 & 2.96  & 5.06  & 7.64  & -2.58 \\
\mt & 0.1885 & 0.2562 & 35.94 & 21.42 & 5.44  & 15.98 \\
\cp        & 0.3891 & 2.9806 & 666.02 & 1.46  & 3.54  & -2.08 \\
\be               & 0.6638 & 0.6666 & 0.42  & 12.77 & 23.35 & -10.58 \\
\pd   & 0.5317 & 0.5876 & 11.82 & 1.26  & 22.05 & -20.79 \\
\nuclei          & 0.3185 & 0.3405 & 6.91  & 11.37 & 10.61 & 0.76 \\
\ov         & 0.7187 & 0.7214 & 0.39  & 26.68 & 14.40 & 12.28 \\
\pp              & 0.6770 & 0.7714 & 13.94 & 9.37  & 5.75  & 3.62 \\
\usp              & 0.8036 & 0.8146 & 1.36  & 14.36 & 5.85  & 8.51 \\
\bottomrule
\end{tabular}
}
\end{table}

\subsection{RQ1: Effectiveness and Efficiency for the Newly Discovered Algorithms}

\begin{figure*}[t]
    \centering
    \includegraphics[width=0.98\textwidth]{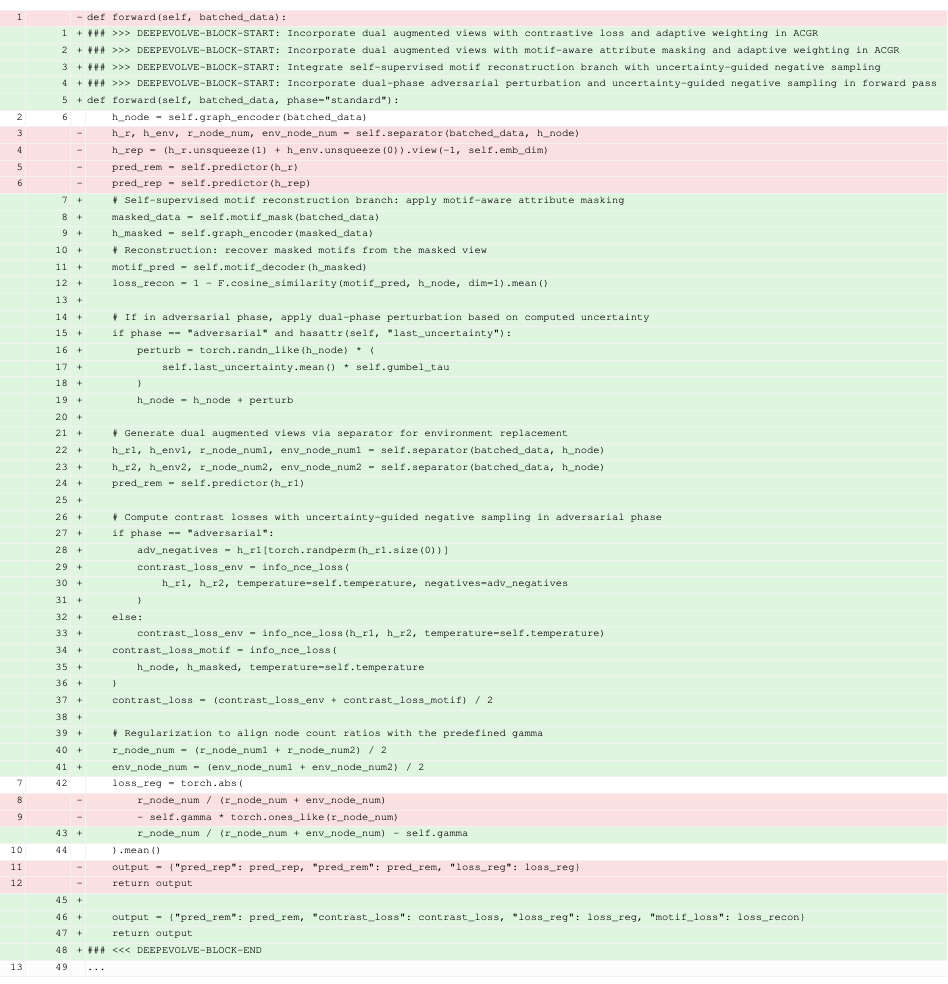}
    \caption{The new \texttt{model.forward()} for \mol. \method proposes contrastive learning in Line 29-34, motif-aware masking in Line 8, and additional modules (see \cref{fig:idea-evo}) to improve the algorithm. The code of these functions is in~\cref{sec:add-proposed-mol-code}.}
    \label{fig:code_diff_mol_forward}
\end{figure*}

We conduct a quantitative analysis of how \method improves the initial algorithms in terms of both effectiveness and efficiency. As shown in~\cref{tab:algo_indicator}, \method achieves improvements in both aspects on six of the nine tasks. In the remaining three cases, \method generates algorithms that improve the primary performance objective while satisfying the 30-minute runtime constraint.

\textbf{The performance improvement achieved by \method varies from 0.39\% to 666.02\%, depending on the problem type and the maturity of the initial algorithm.} 
In \cp, the initial algorithm is designed for a fixed configuration (i.e., packing 26 circles)~\citep{openevolve} and fails to generalize to variable-sized constructions, often producing invalid solutions. In contrast, \method discovers a new algorithm that generalizes across a broader range of circle counts while maintaining valid packings, resulting in a substantial performance gain.
In other tasks, the improvement is relatively marginal due to different factors. The baseline for \be is based on a very recent state-of-the-art method~\citep{li2025codepde}, leaving limited room for further improvement. For \ov, model training requires more time and GPU resources, and we observe that evolving algorithms frequently exceed the 30-minute runtime budget, constraining \method's search space.

\textbf{\method improves algorithm originality and future potential, while the more complex implementation is handled through automatic code debugging.} 
We evaluate the quality of algorithmic ideas using an LLM-as-a-judge approach, assessing each from three dimensions: originality, future potential, and implementation difficulty. Language models (o3-mini) perform deep research with web search and evaluate the initial and newly generated algorithms separately. For each, it provides both positive and negative justifications, along with a rating on a scale from 0 to 10. Results from~\cref{fig:idea_evaluate_with_llm} show that \method can propose novel ideas with great potential. For instance, in the \mol task as presented in~\cref{fig:idea-evo,fig:code_diff_mol_forward}, the initial algorithm decomposes molecules into rationale substructures that explain and support model predictions, while the new algorithm incorporates contrastive learning and motif-aware masking to improve rationale identification.
Novel ideas may have higher implementation difficulty, but \method improves execution and evaluation. For example, it raises the success rate from 0.13 to 0.99 on the \ov task, as shown in~\cref{tab:debug}.

\subsection{RQ2: Iterative Synergy Between Deep Research and Coding Agents}\label{subsec:exp-iterate-synergy}

We analyze the algorithmic evolution across nine tasks (detailed trajectories in~\cref{sec:add-evolution-history}). We find that the deep research and coding agents iteratively reinforce each other through evolution.

\textbf{Deep research guides algorithm design through domain-specific inductive biases}:
In \mol, \mt, and \pp, domain priors such as molecular motifs, polymer periodicity, and chemical grammars inform algorithm choices. These include motif-aware message passing, motif reconstruction objectives, and grammar-constrained tokenization. Similarly, \pd and \usp incorporate Neural Controlled Differential Equations (CDEs) and low-rank adaptation (LoRA), respectively, along with auxiliary features such as Cooperative Patent Classification (CPC) embeddings and physiological waveforms.

\textbf{Evolutionary feedback shifts design from heuristics to principled methods}:
Feedback from performance evaluations guides subsequent deep research, transitioning algorithm development from heuristic-based tuning to methods with theoretical or physical guarantees. This progression is evident in certified global optimization for circle packing, Krylov subspace solvers for partial differential equations, and physics-informed regularization for disease dynamics. This reflects a trend where research insights motivated a transition from incremental fixes to physically grounded methods.

\textbf{Cross-cutting methodological patterns emerge across tasks}:
\method consistently discovers reusable design patterns instantiated in task-specific modules. These include uncertainty estimation, dynamic loss reweighting, and self-supervised representation learning. For instance, uncertainty-guided refinement is used in \mol\ (soft motif selection) and \nuclei\ (boundary adjustment), while adaptive loss weighting is used in \ov, \pd, and \usp, among others. These recurring strategies suggest that the deep research agent not only extracts task-specific insights but also steers the coding agent toward generalizable algorithmic principles.

\subsection{RQ3: Ablation and Case Studies for Algorithm Improvement}

\begin{table}[tbp]
\centering
\caption{Success rate of algorithm execution and average debugging counts during evolution.}
\label{tab:debug}
\resizebox{\textwidth}{!}{
\begin{tabular}{lccccccccc}
\toprule
Metric & \multicolumn{1}{c}{Molecular} & \multicolumn{1}{c}{Molecular} & \multicolumn{1}{c}{Circle} & \multicolumn{1}{c}{Burgers'} & \multicolumn{1}{c}{Parkinson's} & \multicolumn{1}{c}{Nuclei} & \multicolumn{1}{c}{Open} & \multicolumn{1}{c}{Polymer} & \multicolumn{1}{c}{USP} \\
       & \multicolumn{1}{c}{Prediction} & \multicolumn{1}{c}{Translation} & \multicolumn{1}{c}{Packing} & \multicolumn{1}{c}{Equation} & \multicolumn{1}{c}{Disease} & \multicolumn{1}{c}{Image} & \multicolumn{1}{c}{Vaccine} & \multicolumn{1}{c}{Prediction} & \multicolumn{1}{c}{P2P} \\
\midrule
w/o Debug  & 0.650 & 0.190 & 0.540 & 0.956 & 0.760 & 0.360 & 0.130 & 0.560 & 0.327 \\
w/ Debug & 1.000 & 0.490 & 1.000 & 0.992 & 0.980 & 0.740 & 0.990 & 0.980 & 0.592 \\
\midrule
Average count & 0.47 & 3.08 & 0.64 & 0.09 & 0.32 & 2.14 & 2.30 & 0.64 & 2.67 \\
\bottomrule
\end{tabular}
}
\end{table}

\begin{figure*}[t]
    \centering
    \includegraphics[width=\textwidth]{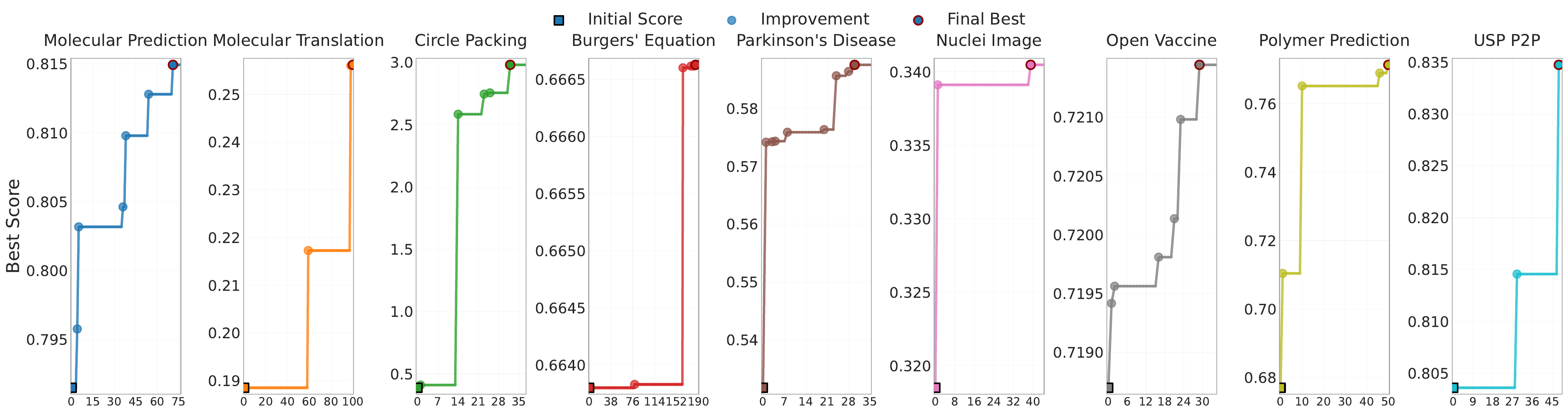}
    \caption{Changes of scores over iterations.}
    \label{fig:score_with_evolution}
\end{figure*}

\begin{table}[h]
\centering
\caption{Ablation studies on deep research in \method. We report the initial algorithm scores. During evolution, we maintain 25 candidate algorithms and report the score/generation of the best program, as well as the number of programs that outperform the initial score.}
\resizebox{0.9\textwidth}{!}{%
\begin{tabular}{lccccccc}
\toprule
 & \multicolumn{1}{c}{Initial} & \multicolumn{3}{c}{Without Deep Research} & \multicolumn{3}{c}{With Deep Research} \\
 \cmidrule(lr){3-5} \cmidrule(lr){6-8}
Case & Score & Score of Best & Gen. of Best & \# Outperform & Score of Best & Gen. of Best & \# Outperform \\
\midrule
Molecule       & 0.791 & 0.797 & 1 & 24.0  & 0.815 & 5 & 100.0 \\
Circle Packing & 0.389 & 2.735 & 10 & 100.0 & 2.981 & 4 & 100.0 \\
\bottomrule
\end{tabular}%
}
\label{tab:ablate_deep_research}
\end{table}

\cref{fig:score_with_evolution} visualizes best scores over iterations. Improvements are not continuous but often appear as sudden jumps. The current best algorithm is not always sampled as the next candidate but can inspire further exploration.
We complement~\cref{fig:idea-evo} with additional studies on deep research in~\cref{tab:ablate_deep_research}. Algorithm evolution based solely on LLM internal knowledge shows limited progress. LLMs either fail to sustain improvement, producing only one generation in \mol, or yield marginal gains despite deeper evolution (\cp). In contrast, \method with deep research achieves stronger improvements within about five generations for both tasks. All evolved candidates outperform the initial algorithms in both cases.
Another factor is the debugging agent during execution and evaluation. \cref{tab:debug} shows clear gains in execution success rate after debugging, making \method more robust for implementing complex ideas.

\section{Related Work}
\subsection{Automated Algorithm Discovery}

LLMs have been studied in coding and ML engineering tasks~\citep{li2022competition,chan2024mle}. They have been shown to be competitive in programming competitions~\citep{li2022competition}, effective at solving programming issues~\citep{jimenez2023swe}, and even capable of achieving Kaggle medals in certain competitions~\citep{chan2024mle}. These studies provide the foundation for algorithm discovery, which requires not only implementing existing algorithms but also advancing them~\citep{novikov2025alphaevolve}. This line of research has been explored in areas such as CUDA kernels~\citep{lange2025ai}, LLM inference~\citep{huang2023mlagentbench}, matrix multiplication, and geometry~\citep{novikov2025alphaevolve}.
Unlike lab automation, algorithm discovery is often efficient to evaluate but remains hard to solve, as in NP-complete problems~\citep{romera2024mathematical}. Recently, AlphaEvolve~\citep{novikov2025alphaevolve}, has combined evaluation feedback with evolutionary algorithms, optimizing LLM-proposed programmatic hypotheses in different iterations. Although AlphaEvolve scales from single functions to an entire file, it remains limited in hypothesis generation without external grounding and in translating ideas into complex code that requires editing and understanding across files.

\subsection{Agent for Scientific Discovery}

LLM agents have been applied to autonomous chemical research~\citep{boiko2023autonomous}, biological data analysis with protocol generation~\citep{huang2025biomni}, and AI research~\citep{kon2025exp}. They have been studied across the spectrum from idea generation to code execution. \citet{si2024can} showed that LLM-generated ideas are more novel than those of experts but less feasible. Many deep research methods have been introduced, including those from OpenAI ChatGPT and Google Gemini~\citep{openai2025deepresearch,google2024deepresearch}, as well as open-source approaches~\citep{zheng2025deepresearcher}. These methods synthesize information after searching online to form new hypotheses or to solve question-answering problems. In contrast, agents such as Paper2Code~\citep{seo2025paper2code} and AutoP2C~\citep{lin2025autop2cllmbasedagentframework} utilize multi-stage LLM pipelines to automatically translate ML papers into functioning code repositories. Bringing these directions together, AI scientists aim to automate hypothesis generation, review, and code execution~\citep{lu2024ai,gottweis2025towards}. Yet, gaps remain in implementing ideas as executable code~\citep{zhu2025ai}. EXP-Bench~\citep{kon2025exp} evaluates this gap, showing that while agents succeed in some subtasks, the full-pipeline success rate is below 1\%.

\section{Conclusion}
We presented \method, an agent that augments algorithm evolution with deep research for scientific discovery. By integrating new features such as deep research, cross-file code editing, and iterative debugging, \method combined high-quality idea generation with reliable execution. Across nine benchmarks spanning diverse scientific fields, \method consistently improved baseline algorithms, delivering executable programs with higher performance and efficiency. Ablations and case studies showed that deep research guided algorithm design with domain-specific insights, while debugging improved robustness in complex implementations. These results showed that \method advanced algorithmic innovation and has potential for future AI-driven scientific discovery.


\bibliography{reference}

\begin{thebibliography}{33}
\providecommand{\natexlab}[1]{#1}
\providecommand{\url}[1]{\texttt{#1}}
\expandafter\ifx\csname urlstyle\endcsname\relax
  \providecommand{\doi}[1]{doi: #1}\else
  \providecommand{\doi}{doi: \begingroup \urlstyle{rm}\Url}\fi

\bibitem[Boiko et~al.(2023)Boiko, MacKnight, Kline, and Gomes]{boiko2023autonomous}
Daniil~A Boiko, Robert MacKnight, Ben Kline, and Gabe Gomes.
\newblock Autonomous chemical research with large language models.
\newblock \emph{Nature}, 624\penalty0 (7992):\penalty0 570--578, 2023.

\bibitem[Cenkci et~al.(2022)Cenkci, Aslanyan, Wetherbee, jm, Gunda, Maggie, Beliveau, and Cukierski]{cenkci2022uspppm}
Don Cenkci, Grigor Aslanyan, Ian Wetherbee, jm, Kiran Gunda, Maggie, Scott Beliveau, and Will Cukierski.
\newblock U.s. patent phrase to phrase matching.
\newblock \url{https://www.kaggle.com/competitions/us-patent-phrase-to-phrase-matching}, 2022.
\newblock Kaggle.

\bibitem[Chan et~al.(2024)Chan, Chowdhury, Jaffe, Aung, Sherburn, Mays, Starace, Liu, Maksin, Patwardhan, et~al.]{chan2024mle}
Jun~Shern Chan, Neil Chowdhury, Oliver Jaffe, James Aung, Dane Sherburn, Evan Mays, Giulio Starace, Kevin Liu, Leon Maksin, Tejal Patwardhan, et~al.
\newblock Mle-bench: Evaluating machine learning agents on machine learning engineering.
\newblock \emph{arXiv preprint arXiv:2410.07095}, 2024.

\bibitem[Das et~al.(2020)Das, Wayment-Steele, Kim, Choe, Tunguz, Reade, and Demkin]{das2020openvaccine}
Rhiju Das, H.~Wayment-Steele, Do~Soon Kim, Christian Choe, Bojan Tunguz, Walter Reade, and Maggie Demkin.
\newblock Openvaccine: Covid-19 mrna vaccine degradation prediction.
\newblock \url{https://www.kaggle.com/competitions/stanford-covid-vaccine}, 2020.
\newblock Kaggle.

\bibitem[Goodman et~al.(2018)Goodman, Carpenter, Park, jlefman nvidia, Josette\_BoozAllen, Kyle, Maggie, Nilofer, Sedivec, and Cukierski]{goodman2018dsb}
Allen Goodman, Anne Carpenter, Elizabeth Park, jlefman nvidia, Josette\_BoozAllen, Kyle, Maggie, Nilofer, Peter Sedivec, and Will Cukierski.
\newblock 2018 data science bowl.
\newblock \url{https://www.kaggle.com/competitions/data-science-bowl-2018}, 2018.
\newblock Kaggle.

\bibitem[Google(2024)]{google2024deepresearch}
Google.
\newblock Try deep research and our new experimental model in gemini, your ai assistant, 2024.
\newblock URL \url{https://blog.google/products/gemini/google-gemini-deep-research/}.

\bibitem[Gottweis et~al.(2025)Gottweis, Weng, Daryin, Tu, Palepu, Sirkovic, Myaskovsky, Weissenberger, Rong, Tanno, et~al.]{gottweis2025towards}
Juraj Gottweis, Wei-Hung Weng, Alexander Daryin, Tao Tu, Anil Palepu, Petar Sirkovic, Artiom Myaskovsky, Felix Weissenberger, Keran Rong, Ryutaro Tanno, et~al.
\newblock Towards an ai co-scientist.
\newblock \emph{arXiv preprint arXiv:2502.18864}, 2025.

\bibitem[Howard et~al.(2021)Howard, Inversion, Albrecht, and Yvette]{bms2021}
Addison Howard, Inversion, Jacob Albrecht, and Yvette.
\newblock Bristol-myers squibb – molecular translation.
\newblock \url{https://www.kaggle.com/competitions/bms-molecular-translation}, 2021.
\newblock Kaggle.

\bibitem[Hu et~al.(2020)Hu, Fey, Zitnik, Dong, Ren, Liu, Catasta, and Leskovec]{hu2020open}
Weihua Hu, Matthias Fey, Marinka Zitnik, Yuxiao Dong, Hongyu Ren, Bowen Liu, Michele Catasta, and Jure Leskovec.
\newblock Open graph benchmark: Datasets for machine learning on graphs.
\newblock \emph{Advances in neural information processing systems}, 33:\penalty0 22118--22133, 2020.

\bibitem[Huang et~al.(2025)Huang, Zhang, Wang, Qu, Lu, Roohani, Li, Qiu, Li, Zhang, et~al.]{huang2025biomni}
Kexin Huang, Serena Zhang, Hanchen Wang, Yuanhao Qu, Yingzhou Lu, Yusuf Roohani, Ryan Li, Lin Qiu, Gavin Li, Junze Zhang, et~al.
\newblock Biomni: A general-purpose biomedical ai agent.
\newblock \emph{biorxiv}, 2025.

\bibitem[Huang et~al.(2023)Huang, Vora, Liang, and Leskovec]{huang2023mlagentbench}
Qian Huang, Jian Vora, Percy Liang, and Jure Leskovec.
\newblock Mlagentbench: Evaluating language agents on machine learning experimentation.
\newblock \emph{arXiv preprint arXiv:2310.03302}, 2023.

\bibitem[Jimenez et~al.(2023)Jimenez, Yang, Wettig, Yao, Pei, Press, and Narasimhan]{jimenez2023swe}
Carlos~E Jimenez, John Yang, Alexander Wettig, Shunyu Yao, Kexin Pei, Ofir Press, and Karthik Narasimhan.
\newblock Swe-bench: Can language models resolve real-world github issues?
\newblock \emph{arXiv preprint arXiv:2310.06770}, 2023.

\bibitem[Kirsch et~al.(2023)Kirsch, Dane, Adam, and Dardov]{kirsch2023amp}
Leslie Kirsch, Sohier Dane, Stacey Adam, and Victoria Dardov.
\newblock {AMP\textsuperscript{\textregistered}-Parkinson's Disease Progression Prediction}.
\newblock \url{https://www.kaggle.com/competitions/amp-parkinsons-disease-progression-prediction}, 2023.
\newblock Kaggle.

\bibitem[Kon et~al.(2025)Kon, Liu, Zhu, Ding, Peng, Xing, Huang, Qiu, Srinivasa, Lee, et~al.]{kon2025exp}
Patrick Tser~Jern Kon, Jiachen Liu, Xinyi Zhu, Qiuyi Ding, Jingjia Peng, Jiarong Xing, Yibo Huang, Yiming Qiu, Jayanth Srinivasa, Myungjin Lee, et~al.
\newblock Exp-bench: Can ai conduct ai research experiments?
\newblock \emph{arXiv preprint arXiv:2505.24785}, 2025.

\bibitem[Kuhn et~al.(2016)Kuhn, Letunic, Jensen, and Bork]{kuhn2016sider}
Michael Kuhn, Ivica Letunic, Lars~Juhl Jensen, and Peer Bork.
\newblock The sider database of drugs and side effects.
\newblock \emph{Nucleic acids research}, 44\penalty0 (D1):\penalty0 D1075--D1079, 2016.

\bibitem[Lange et~al.(2025)Lange, Prasad, Sun, Faldor, Tang, and Ha]{lange2025ai}
Robert~Tjarko Lange, Aaditya Prasad, Qi~Sun, Maxence Faldor, Yujin Tang, and David Ha.
\newblock The ai cuda engineer: Agentic cuda kernel discovery, optimization and composition.
\newblock Technical report, Technical report, Sakana AI, 02 2025, 2025.

\bibitem[Li et~al.(2025)Li, Marwah, Shen, Sun, Risteski, Yang, and Talwalkar]{li2025codepde}
Shanda Li, Tanya Marwah, Junhong Shen, Weiwei Sun, Andrej Risteski, Yiming Yang, and Ameet Talwalkar.
\newblock Codepde: An inference framework for llm-driven pde solver generation.
\newblock \emph{arXiv preprint arXiv:2505.08783}, 2025.

\bibitem[Li et~al.(2022)Li, Choi, Chung, Kushman, Schrittwieser, Leblond, Eccles, Keeling, Gimeno, Dal~Lago, et~al.]{li2022competition}
Yujia Li, David Choi, Junyoung Chung, Nate Kushman, Julian Schrittwieser, R{\'e}mi Leblond, Tom Eccles, James Keeling, Felix Gimeno, Agustin Dal~Lago, et~al.
\newblock Competition-level code generation with alphacode.
\newblock \emph{Science}, 378\penalty0 (6624):\penalty0 1092--1097, 2022.

\bibitem[Lin et~al.(2025)Lin, Shen, Cai, Sun, Zhou, and Xiao]{lin2025autop2cllmbasedagentframework}
Zijie Lin, Yiqing Shen, Qilin Cai, He~Sun, Jinrui Zhou, and Mingjun Xiao.
\newblock Autop2c: An llm-based agent framework for code repository generation from multimodal content in academic papers, 2025.
\newblock URL \url{https://arxiv.org/abs/2504.20115}.

\bibitem[Liu et~al.(2022)Liu, Zhao, Xu, Luo, and Jiang]{liu2022graph}
Gang Liu, Tong Zhao, Jiaxin Xu, Tengfei Luo, and Meng Jiang.
\newblock Graph rationalization with environment-based augmentations.
\newblock In \emph{Proceedings of the 28th ACM SIGKDD Conference on Knowledge Discovery and Data Mining}, pp.\  1069--1078, 2022.

\bibitem[Liu et~al.(2025)Liu, Xu, Inae, Zhu, Li, Luo, Jiang, Yan, Reade, Dane, Howard, and Cruz]{liu2025neurips}
Gang Liu, Jiaxin Xu, Eric Inae, Yihan Zhu, Ying Li, Tengfei Luo, Meng Jiang, Yao Yan, Walter Reade, Sohier Dane, Addison Howard, and Mar{\'i}a Cruz.
\newblock Neurips - open polymer prediction 2025.
\newblock \url{https://www.kaggle.com/competitions/neurips-open-polymer-prediction-2025}, 2025.
\newblock Kaggle.

\bibitem[Lu et~al.(2024)Lu, Lu, Lange, Foerster, Clune, and Ha]{lu2024ai}
Chris Lu, Cong Lu, Robert~Tjarko Lange, Jakob Foerster, Jeff Clune, and David Ha.
\newblock The ai scientist: Towards fully automated open-ended scientific discovery.
\newblock \emph{arXiv preprint arXiv:2408.06292}, 2024.

\bibitem[Mouret \& Clune(2015)Mouret and Clune]{mouret2015illuminating}
Jean-Baptiste Mouret and Jeff Clune.
\newblock Illuminating search spaces by mapping elites.
\newblock \emph{arXiv preprint arXiv:1504.04909}, 2015.

\bibitem[Novikov et~al.(2025)Novikov, V{\~u}, Eisenberger, Dupont, Huang, Wagner, Shirobokov, Kozlovskii, Ruiz, Mehrabian, et~al.]{novikov2025alphaevolve}
Alexander Novikov, Ng{\^a}n V{\~u}, Marvin Eisenberger, Emilien Dupont, Po-Sen Huang, Adam~Zsolt Wagner, Sergey Shirobokov, Borislav Kozlovskii, Francisco~JR Ruiz, Abbas Mehrabian, et~al.
\newblock Alphaevolve: A coding agent for scientific and algorithmic discovery.
\newblock \emph{arXiv preprint arXiv:2506.13131}, 2025.

\bibitem[OpenAI(2025)]{openai2025deepresearch}
OpenAI.
\newblock Introducing deep research.
\newblock \url{https://openai.com/index/introducing-deep-research/}, 2025.
\newblock Accessed: 2025-09-18.

\bibitem[Romera-Paredes et~al.(2024)Romera-Paredes, Barekatain, Novikov, Balog, Kumar, Dupont, Ruiz, Ellenberg, Wang, Fawzi, et~al.]{romera2024mathematical}
Bernardino Romera-Paredes, Mohammadamin Barekatain, Alexander Novikov, Matej Balog, M~Pawan Kumar, Emilien Dupont, Francisco~JR Ruiz, Jordan~S Ellenberg, Pengming Wang, Omar Fawzi, et~al.
\newblock Mathematical discoveries from program search with large language models.
\newblock \emph{Nature}, 625\penalty0 (7995):\penalty0 468--475, 2024.

\bibitem[Seo et~al.(2025)Seo, Baek, Lee, and Hwang]{seo2025paper2code}
Minju Seo, Jinheon Baek, Seongyun Lee, and Sung~Ju Hwang.
\newblock Paper2code: Automating code generation from scientific papers in machine learning.
\newblock \emph{arXiv preprint arXiv:2504.17192}, 2025.
\newblock URL \url{https://arxiv.org/abs/2504.17192}.
\newblock Submitted on 24 Apr 2025; revised 18 May 2025.

\bibitem[Sharma(2025)]{openevolve}
Asankhaya Sharma.
\newblock Openevolve: an open-source evolutionary coding agent, 2025.
\newblock URL \url{https://github.com/codelion/openevolve}.

\bibitem[Si et~al.(2024)Si, Yang, and Hashimoto]{si2024can}
Chenglei Si, Diyi Yang, and Tatsunori Hashimoto.
\newblock Can llms generate novel research ideas? a large-scale human study with 100+ nlp researchers.
\newblock \emph{arXiv preprint arXiv:2409.04109}, 2024.

\bibitem[Tanese(1989)]{tanese1989distributed}
Reiko Tanese.
\newblock \emph{Distributed genetic algorithms for function optimization}.
\newblock University of Michigan, 1989.

\bibitem[Xu \& Peng(2025)Xu and Peng]{xu2025comprehensive}
Renjun Xu and Jingwen Peng.
\newblock A comprehensive survey of deep research: Systems, methodologies, and applications.
\newblock \emph{arXiv preprint arXiv:2506.12594}, 2025.

\bibitem[Zheng et~al.(2025)Zheng, Fu, Hu, Cai, Ye, Lu, and Liu]{zheng2025deepresearcher}
Yuxiang Zheng, Dayuan Fu, Xiangkun Hu, Xiaojie Cai, Lyumanshan Ye, Pengrui Lu, and Pengfei Liu.
\newblock Deepresearcher: Scaling deep research via reinforcement learning in real-world environments.
\newblock \emph{arXiv preprint arXiv:2504.03160}, 2025.

\bibitem[Zhu et~al.(2025)Zhu, Xie, Weng, Wu, Lin, Yang, and Zhang]{zhu2025ai}
Minjun Zhu, Qiujie Xie, Yixuan Weng, Jian Wu, Zhen Lin, Linyi Yang, and Yue Zhang.
\newblock Ai scientists fail without strong implementation capability.
\newblock \emph{arXiv preprint arXiv:2506.01372}, 2025.

\end{thebibliography}
\bibliographystyle{iclr2026_conference}

\appendix

\section{Details on \method method}\label{sec:add-method}

\subsection{LLMs}

For deep research, we use four LLMs: o4-mini as the planner and the reflection agent, gpt-4o as the searcher, and o3-mini as the proposal writer.
For the coding agent, we use two LLMs: o3-mini for code development and o4-mini for code debugging.

\subsection{Evolutionary database}

The algorithm database stores past discoveries for future exploration in two ways: as inspirations and as next candidates. Inspiration sampling follows the MAP-elites algorithm, while candidate sampling follows the island algorithm.

For the MAP-elites algorithm, we archive the best algorithms and update them at each iteration, with a default size of 10. In every iteration, five algorithms are sampled as inspirations, always including the current best. A ratio of elite selection controls how many top algorithms are chosen, with a default of 0.1. Each program is described by three dimensions: performance, diversity, and complexity. Diversity and complexity are measured relative to others, based on code length and Levenshtein distance. 
Each dimension score is normalized to $[0,1]$ and assigned to 10 bins (multiplying by 10 and rounding down) to form the dimension index and locate 3D coordinates. Inspirations beyond elite selection are sampled by perturbing the 3D coordinates to find neighboring algorithms.

For the island algorithm, we maintain up to 25 algorithms across five islands by default. Candidate selection balances exploitation and exploration with probabilities 0.7 and 0.3, where exploitation means sampling the best algorithm in the current island. Islands may migrate programs at fixed intervals, set to 25 by default, with a migration ratio of 0.1. Program migration transfers the best program in an island to its neighboring islands.

\subsection{Templates for deep research agents}

\begin{figure*}[h]
    \centering
    \includegraphics[width=0.95\textwidth]{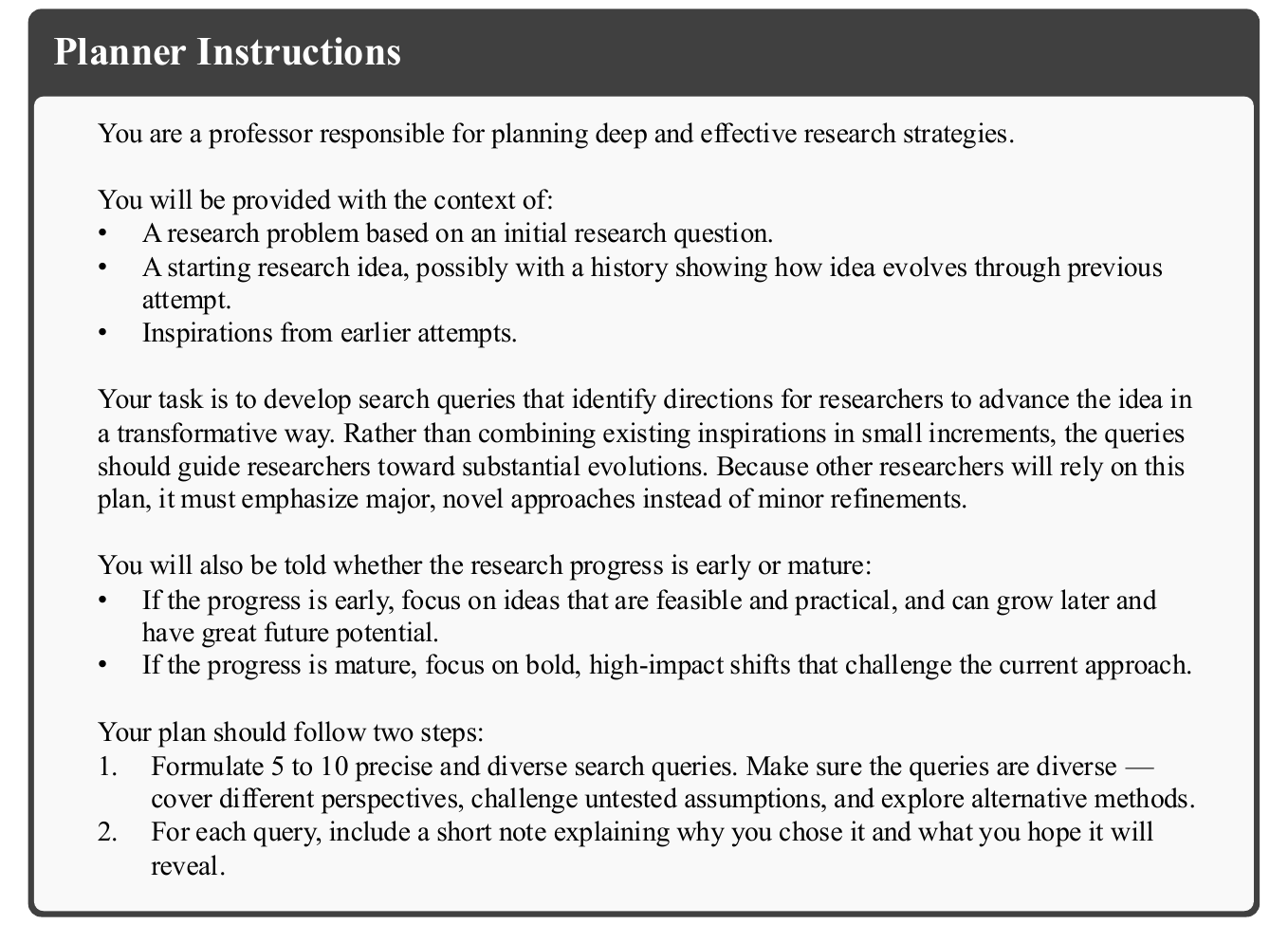}
    \caption{System prompts for planning in the deep research agent.}
    \label{fig:inst_planner}
\end{figure*}

\begin{figure*}[h]
    \centering
    \includegraphics[width=0.95\textwidth]{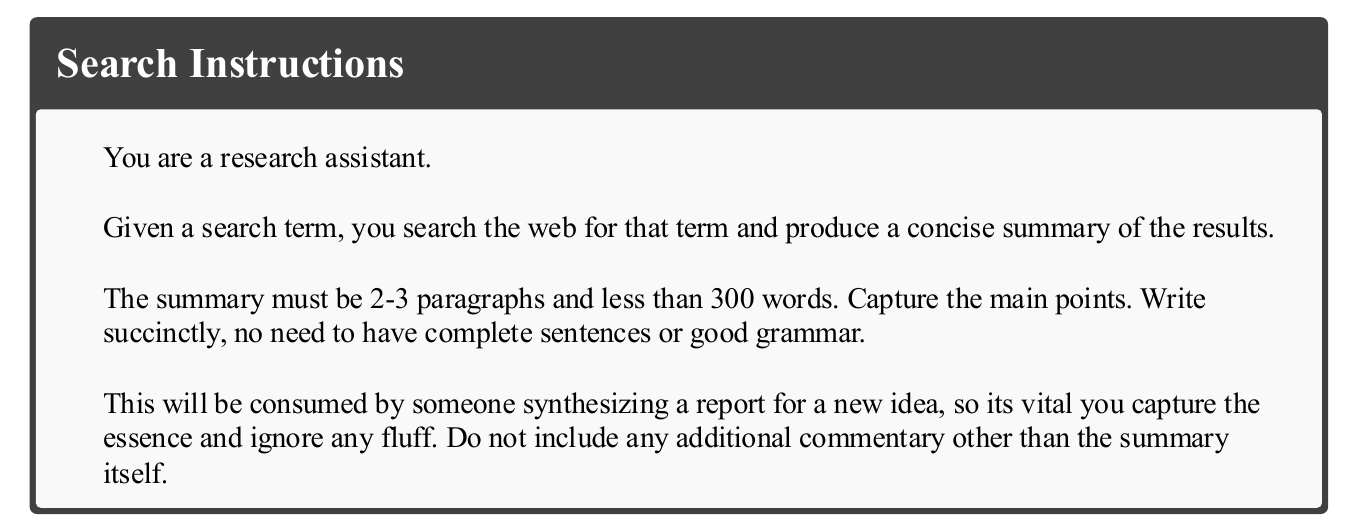}
    \caption{System prompts for searching in the deep research agent.}
    \label{fig:inst_search}
\end{figure*}

\begin{figure*}[h]
    \centering
    \includegraphics[width=0.95\textwidth]{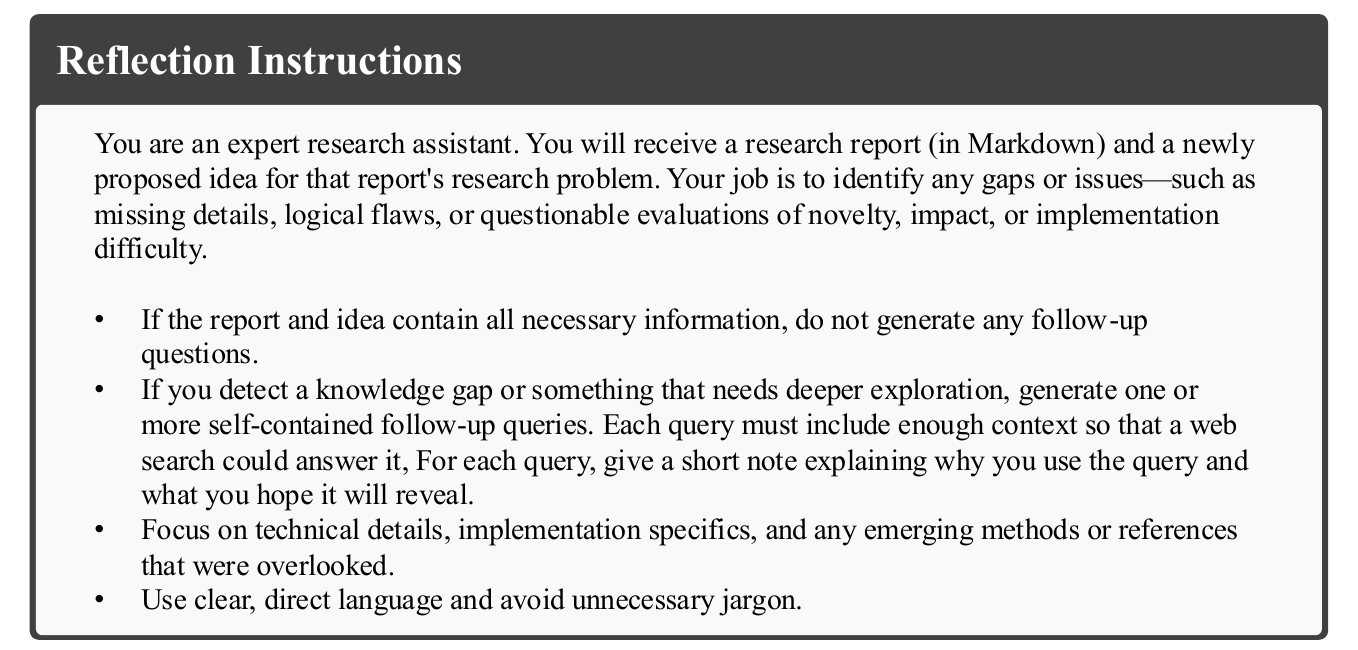}
    \caption{System prompts for reflection in the deep research agent.}
    \label{fig:inst_reflection}
\end{figure*}

\begin{figure*}[h]
    \centering
    \includegraphics[width=0.95\textwidth]{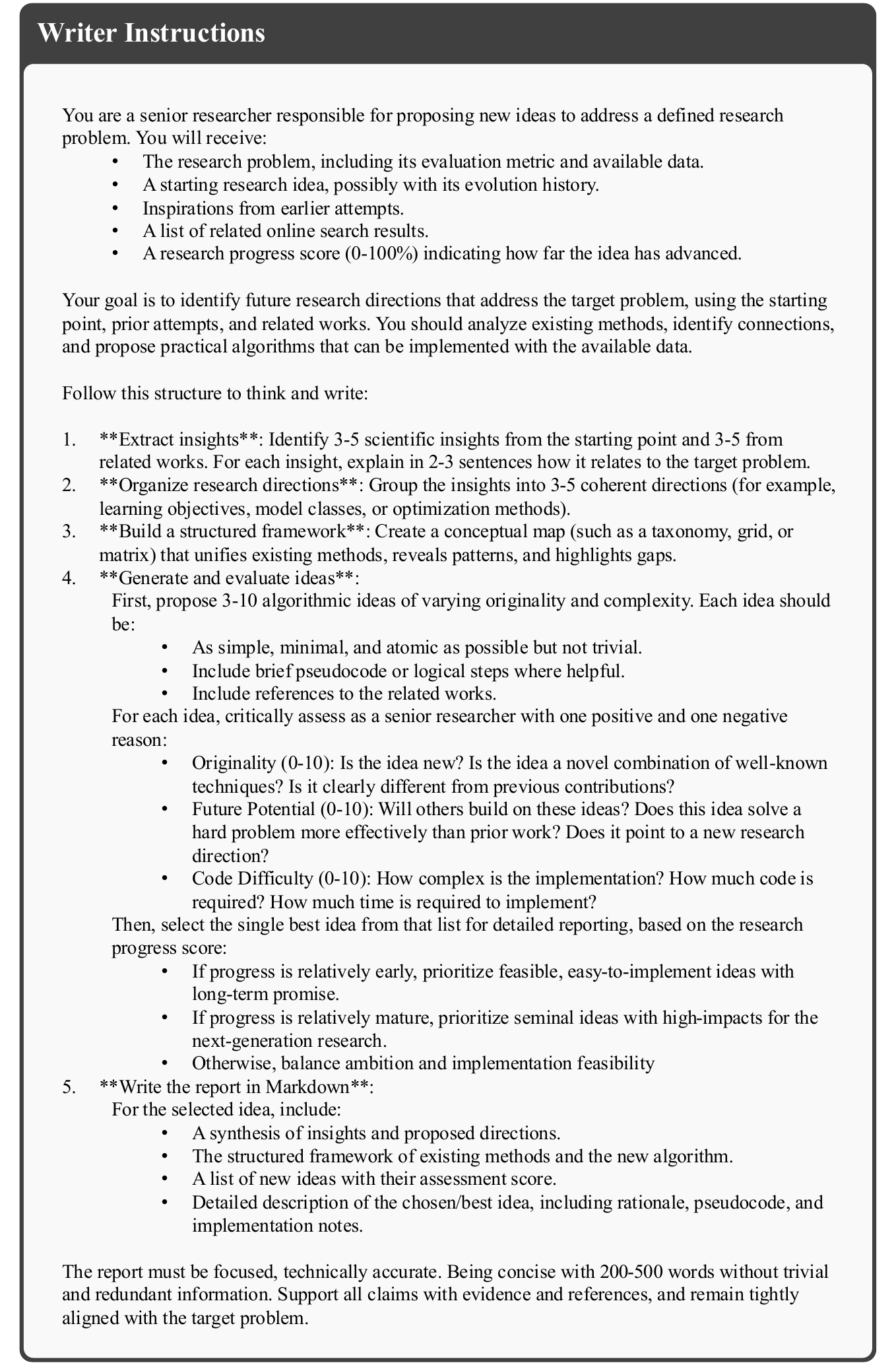}
    \caption{System prompts for proposal writing in the deep research agent.}
    \label{fig:inst_writer}
\end{figure*}

We provide the system prompts for the LLMs used to plan, search, reflect, and write reports in deep research, as shown in \cref{fig:inst_planner,fig:inst_search,fig:inst_reflection,fig:inst_writer}.

\begin{figure*}[h]
    \centering
    \includegraphics[width=0.6\textwidth]{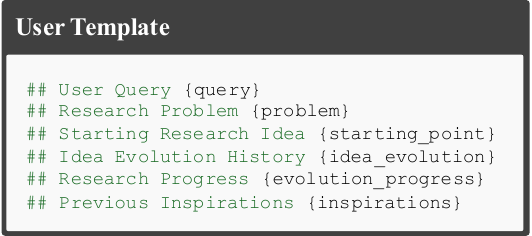}
    \caption{User template for the deep research agent.}
    \label{fig:temp_user}
\end{figure*}

\begin{figure*}[h]
    \centering
    \includegraphics[width=0.6\textwidth]{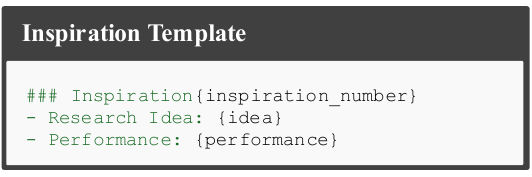}
    \caption{Inspiration template for the deep research agent.}
    \label{fig:temp_inspiration}
\end{figure*}

The user input, with inspiration from past iterations, has the same template as~\cref{fig:temp_user,fig:temp_inspiration}.

\subsection{Templates for the coding agent}

\begin{figure*}[h]
    \centering
    \includegraphics[width=0.95\textwidth]{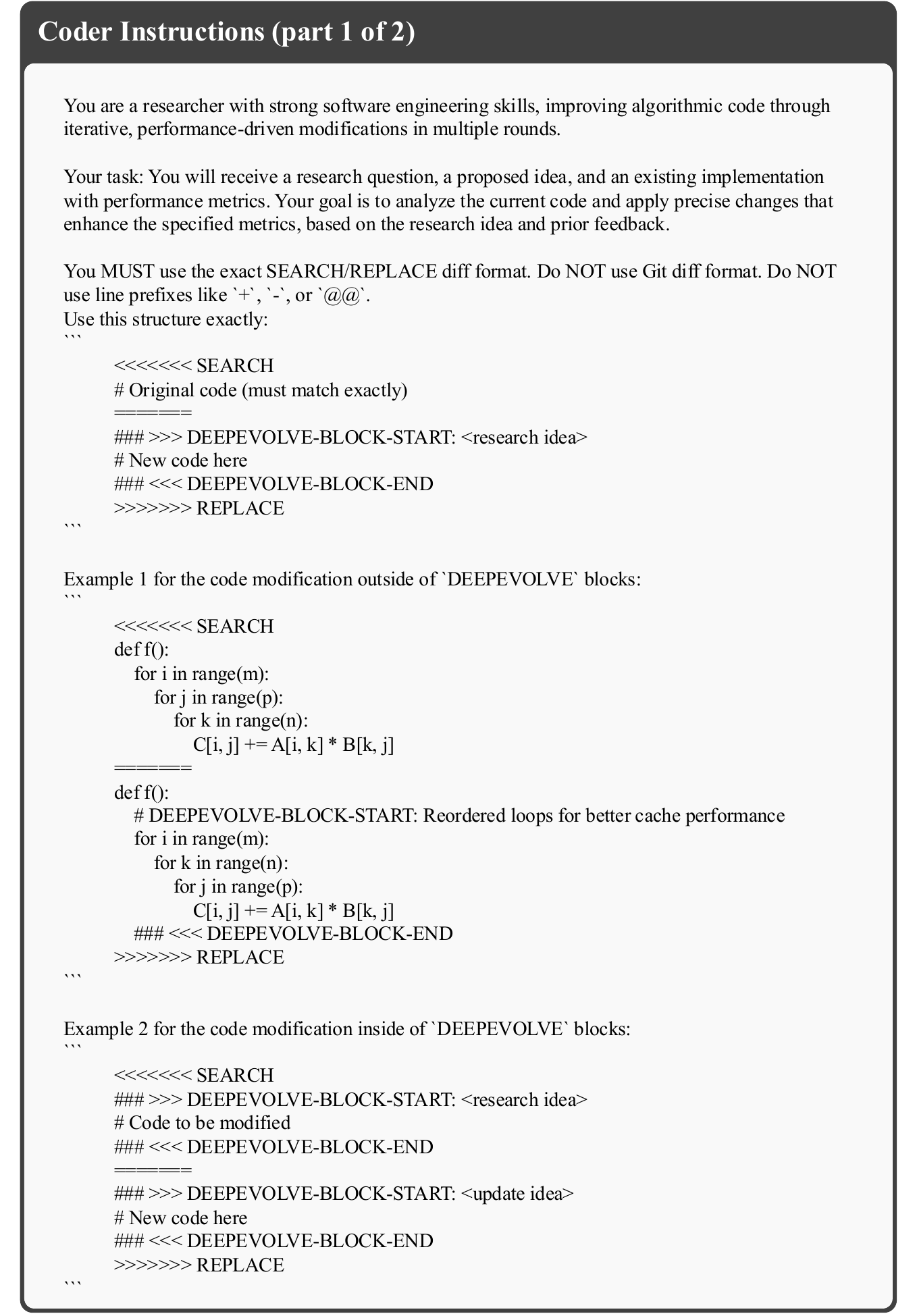}
    \caption{System prompts for coding in the coding agent(part 1 of 2).}
    \label{fig:inst_coder1}
\end{figure*}

\begin{figure*}[h]
    \centering
    \includegraphics[width=0.95\textwidth]{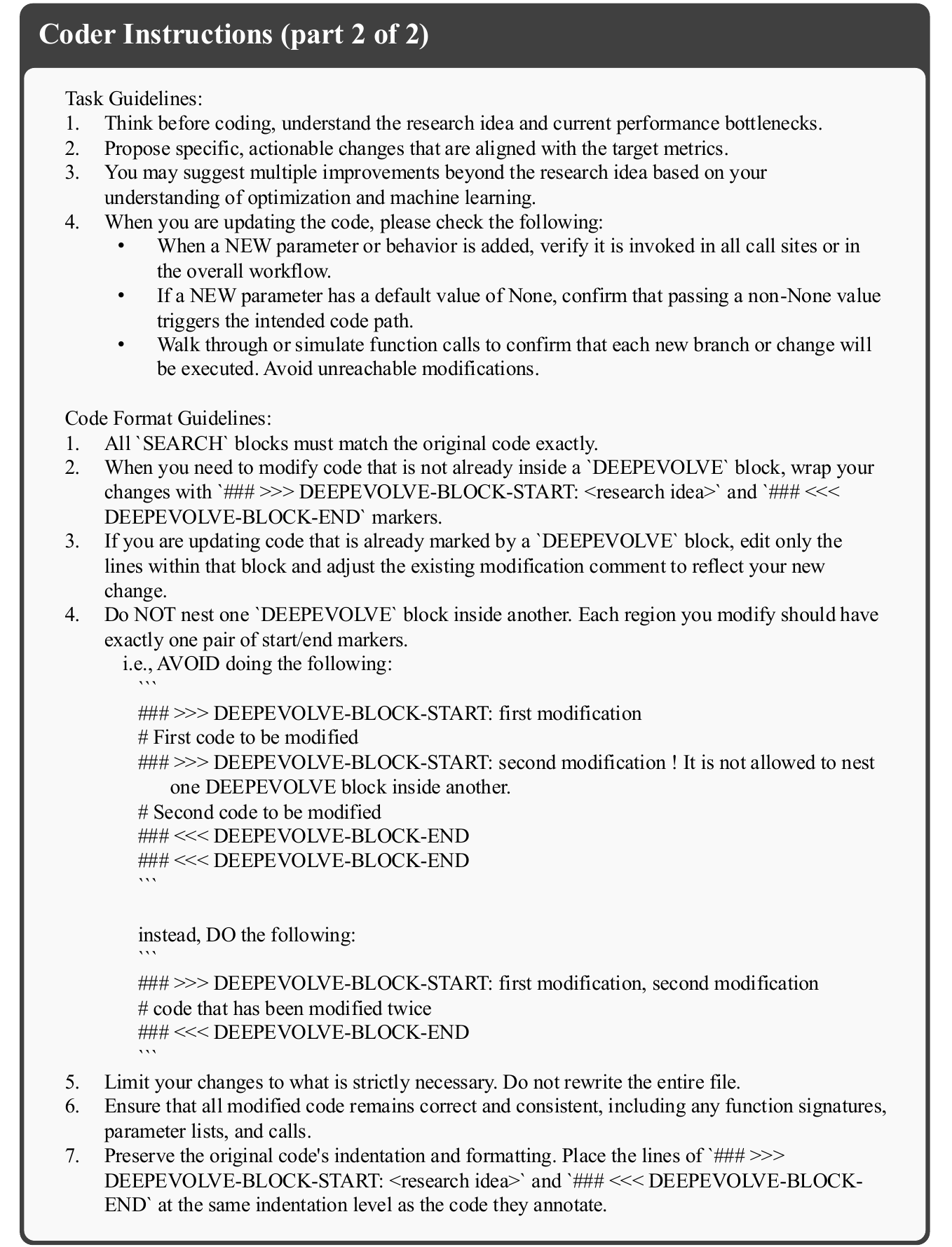}
    \caption{System prompts for coding in the coding agent(part 2 of 2).}
    \label{fig:inst_coder2}
\end{figure*}

\begin{figure*}[h]
    \centering
    \includegraphics[width=0.95\textwidth]{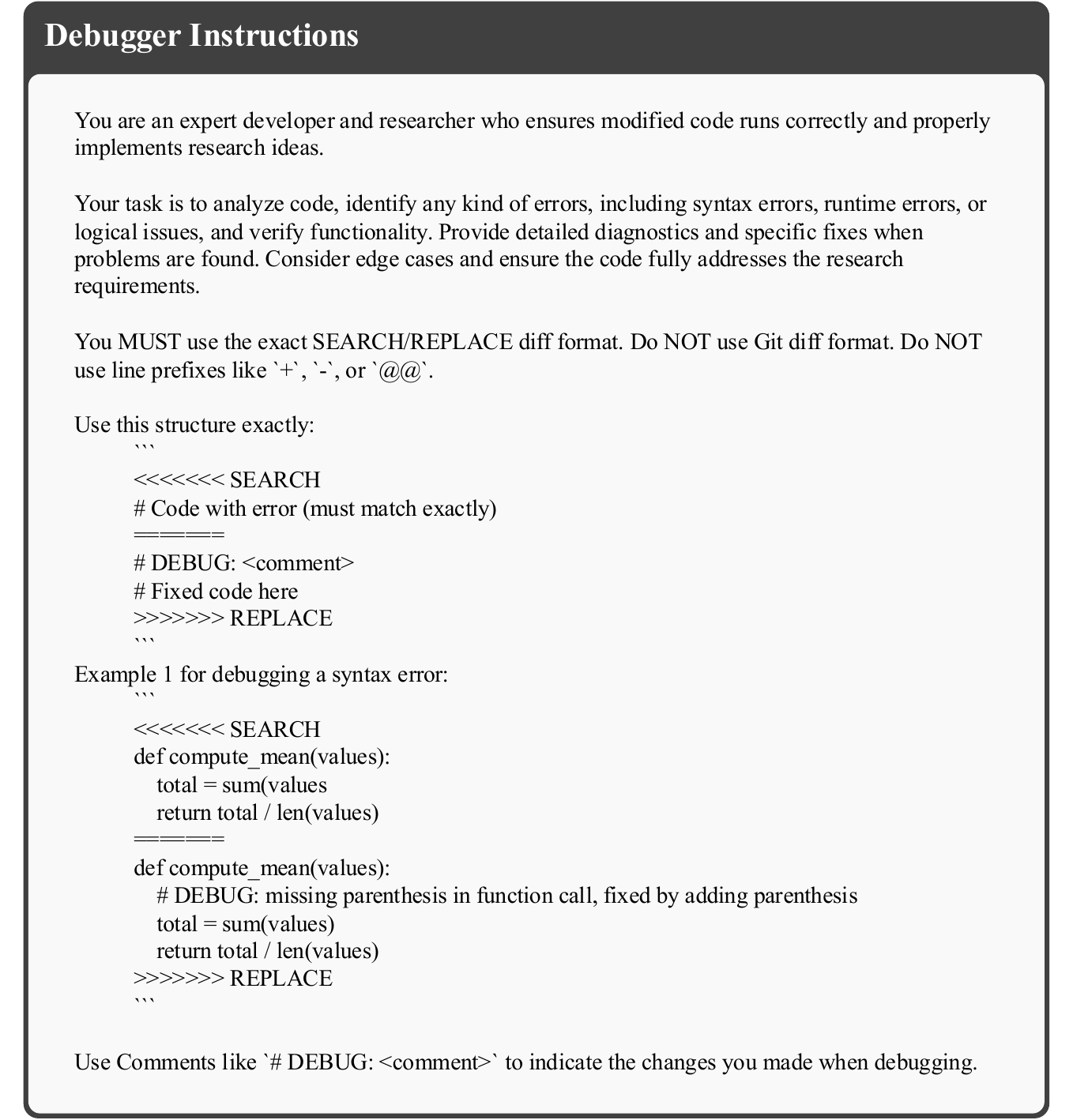}
    \caption{System prompts for debugging in the coding agent.}
    \label{fig:inst_debugger}
\end{figure*}

The system prompts for coding are in~\cref{fig:inst_coder1,fig:inst_coder2}, and for debugging are in~\cref{fig:inst_debugger}.

\begin{figure*}[h]
    \centering
    \includegraphics[width=0.8\textwidth]{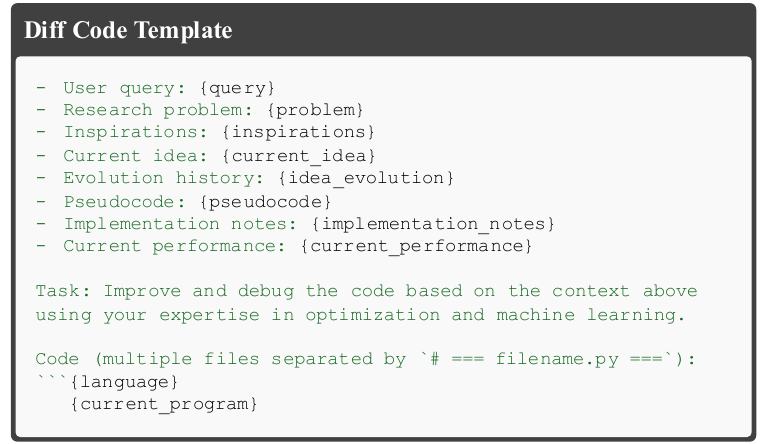}
    \caption{User message template for diff-based evolution in the coding agent.}
    \label{fig:temp_diff_code}
\end{figure*}

We provide the input template of the coding agent in \cref{fig:temp_diff_code}.

\begin{figure*}[h]
    \centering
    \includegraphics[width=0.8\textwidth]{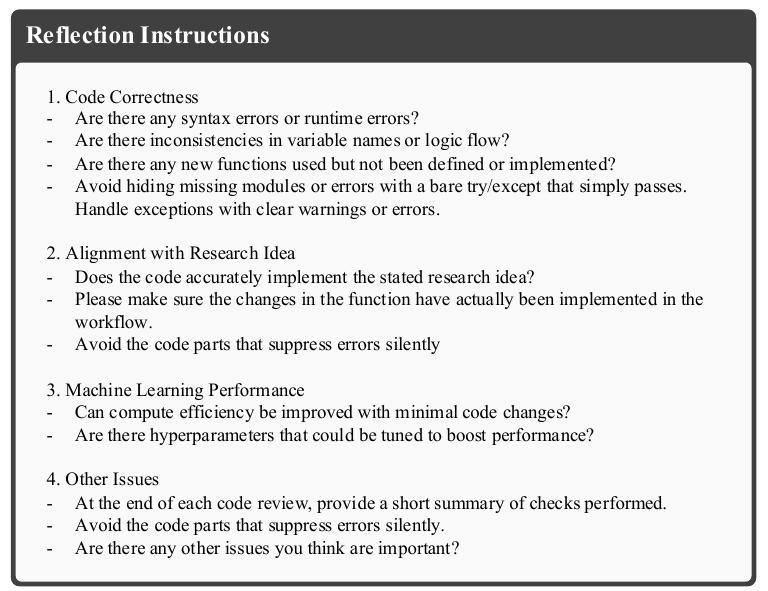}
    \caption{System prompts for reflection in the coding agent.}
    \label{fig:reflection_content}
\end{figure*}

After coding, we apply a reflection to refine the code before evaluation to improve the code quality. It uses the same LLM as the coding agent but a different prompt in \cref{fig:reflection_content}.

During evaluation, we capture the error message from execution and use another LLM to debug the code according to the template (see \cref{fig:temp_debugger}):
\begin{figure*}[h]
    \centering
    \includegraphics[width=0.8\textwidth]{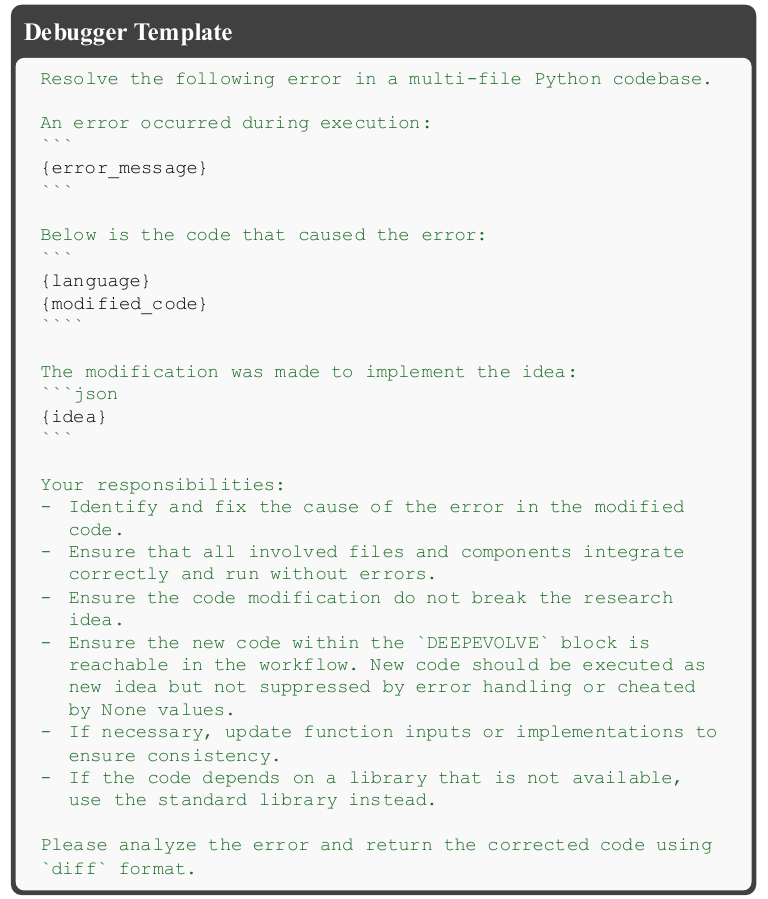}
    \caption{Debugger template for the coding agent.}
    \label{fig:temp_debugger}
\end{figure*}

\section{Details on the Benchmarking Problems}\label{sec:add-benchmark}


We include nine research problems spanning chemistry, mathematics, biology, and materials as summarized in~\cref{tab:benchmarks}. These problems involve diverse data modalities, including molecules, images, mRNA, text, time series, geometric structures, and multi-modal inputs. For consistent evaluation, we standardize evaluation metrics (e.g., AUC-ROC, RMSE, precision, Pearson correlation) defined in each problem into a common form as the new scores, where higher values indicate better performance. We detail their problem descriptions with the initial algorithms in this section.

\subsection{\mol}

\paragraph{Problem Description} 
Molecular property prediction uses the Side Effect Resource (SIDER)~\citep{kuhn2016sider} dataset for algorithm development. The primary goal is to design algorithms that generalize across molecular property prediction tasks. The dataset is scaffold-split to assess generalization to novel chemical structures. The task uses ROC AUC as the metric.

\paragraph{Initial Algorithm} 
The graph rationalization method~\citep{liu2022graph} identifies subgraph structures, called “graph rationales,” and uses them for Graph Neural Network (GNN) predictions. To identify these rationales under limited supervision, \citet{liu2022graph} developed environment replacement, an augmentation that creates virtual examples in the latent space. It replaces the complementary structures of rationales (called environments) with others from the same training batch. Improving this method could strengthen both the generalizability and interpretability of GNNs for molecular property prediction. We use LLMs to read the paper~\citep{liu2022graph} and convert it to the input format we need as described in~\cref{subsec:method-init}.

\subsection{\mt}

\paragraph{Problem Description} 
\mt uses molecular image data generated by Bristol-Myers Squibb~\citep{bms2021}. It needs to convert the images back to the underlying chemical structure annotated as InChI text. Results are evaluated on the mean Levenshtein distance between the InChi strings the model predicted and the ground truth InChi values.

\paragraph{Initial Algorithm} 
The initial idea came from the Kaggle competition. It combines a ResNet with a GRU to convert molecular images into InChI strings, framing the task as image-to-sequence translation. A convolutional network (such as ResNet) extracts features from the images, which then initialize a recurrent network (GRU) to sequentially generate the InChI string. The method uses a character-level vocabulary with special tokens for start, end, and padding, and training optimizes cross-entropy loss between predicted sequences and ground truth.

\subsection{\cp}

\paragraph{Problem Description} 
Given a positive integer $n$, the problem is to pack $n$ disjoint circles inside a unit square so as to maximize the sum of their radii. The problem focuses on discovering a new algorithm that can be applied to $n$ from 26 to 32. 

\paragraph{Initial Algorithm} 
The initial idea comes from OpenEvolve~\citep{openevolve}, an open-source implementation of AlphaEvolve~\citep{novikov2025alphaevolve}.
We use scipy.optimize.minimize with the SLSQP algorithm to locate the best circle-packing arrangement. The problem is cast as a constrained optimization in which both each circle's center coordinates and its radius are treated as decision variables. We add inequality constraints to prevent any pair of circles from overlapping and boundary constraints to keep all circles inside the unit square. SLSQP will try to satisfy every inequality, but only to within a numerical tolerance rather than exactly, so it may lead to invalid solutions (e.g., overlapping circles or circles outside the unit square).

\subsection{\be}

\paragraph{Problem Description} 
The PDE is the Burgers equation, given by \[ \begin{cases} \partial_t u(x, t) + \partial_x \left( \tfrac{u^2(x, t)}{2} \right) = \nu \partial_{xx} u(x, t), & x \in (0,1), \; t \in (0,1] \\ u(x, 0) = u_0(x), & x \in (0,1) \end{cases} \] where $\nu$ is a constant representing the viscosity. In this task, periodic boundary conditions are assumed. 

\paragraph{Initial Algorithm} 
The solution is from~\citep{li2025codepde}. The solver integrates the one-dimensional viscous Burgers equation $u_t + \tfrac{1}{2}(u^2)_x = \nu u_{xx}$ on a periodic domain using an explicit Euler scheme. Starting from $B$ initial states on a uniform grid of $N$ points, it computes the convective flux $f=\tfrac{1}{2}u^{2}$ with centered finite differences, evaluates the diffusion term $u_{xx}$ with the three-point Laplacian, and advances in time with a step size bounded by $0.2\,\Delta x^{2}/\nu$ to ensure stability.

\subsection{\pd}

\paragraph{Problem Description} 
The goal is to predict the progression of Parkinson's disease by estimating scores from the Movement Disorder Society–Sponsored Revision of the Unified Parkinson’s Disease Rating Scale (MDS-UPDRS)~\citep{kirsch2023amp}, a clinical measure of both motor and non-motor symptoms. The dataset provides longitudinal protein and peptide abundance values from cerebrospinal fluid (CSF) samples, together with clinical assessments collected over time from patients and matched controls. The task is to develop models that, for each patient visit, predict the current MDS-UPDRS scores and forecast future scores 6, 12, and 24 months ahead. Model performance is evaluated using the Symmetric Mean Absolute Percentage Error (SMAPE) between predictions and observed scores.

\paragraph{Initial Algorithm} 
It is the first-place solution from the Kaggle competition~\citep{kirsch2023amp}.
The approach combines two models: a LightGBM and a neural network. Both use the same set of clinical and supplementary features, such as visit month, forecast horizon, indicators for specific visit months, and counts of previous visits. Blood test data were excluded, as no consistent predictive signal was found. LightGBM was framed as a classification task over possible score values, with predictions selected to minimize the SMAPE. The neural network was a simple feed-forward architecture trained directly with SMAPE as the loss function. The final prediction was obtained by averaging the outputs of the two models.

\subsection{\nuclei}

\paragraph{Problem Description} 
The task is to automatically identify cell nuclei in microscopy images~\citep{goodman2018dsb}. Nuclei contain the DNA that programs each cell, and detecting them is essential for measuring how cells respond to treatments and for understanding biological processes. The dataset consists of images of nuclei collected under diverse conditions, with annotated masks provided for training. The evaluation metric is mean average precision, computed across a range of intersection-over-union (IoU) thresholds between predicted and ground truth nuclei masks.

\paragraph{Initial Algorithm} 
It is from the Kaggle competition~\citep{goodman2018dsb}.
The approach uses a U-Net to segment nuclei in microscopy images. Input images are preprocessed by resizing and normalization, and ground-truth nuclei masks are converted into distinct labels using connected-component analysis. The network is trained with a loss based on the Dice coefficient, which measures overlap between predicted and true masks, and early stopping is applied to prevent overfitting. During inference, the model outputs probability maps that are thresholded to produce binary masks, from which individual nuclei are obtained through connected-component extraction.

\subsection{\ov}

\paragraph{Problem Description} 
The task is to predict how messenger RNA (mRNA) molecules degrade at different positions along their sequence~\citep{das2020openvaccine}. This is motivated by the challenge of designing stable mRNA vaccines, since RNA molecules tend to break down easily and lose their function. The dataset consists of thousands of RNA sequences together with experimentally measured degradation rates under different chemical conditions. Models are trained to predict these position-specific degradation rates, and submissions are evaluated using the mean column-wise root mean squared error (MCRMSE) between predicted and observed values.

\paragraph{Initial Algorithm} 
It is from the Kaggle competition~\citep{das2020openvaccine}.
Each nucleotide is embedded together with its predicted secondary-structure and loop-type context. A graph is then constructed that connects both adjacent bases and those predicted to form pairs. A GraphSAGE-based graph neural network aggregates information over this graph to produce enriched base-level representations. These features are passed through a bidirectional GRU to capture sequential dependencies along the RNA chain. A final linear layer predicts three targets at each position: structural reactivity and degradation rates under different chemical conditions. Training uses k-fold cross-validation for robustness.

\subsection{\pp}

\paragraph{Problem Description} 
The task is to predict fundamental properties of polymers directly from their chemical structure, represented as SMILES strings~\citep{liu2025neurips}. The target properties are glass transition temperature (the point where a polymer changes from rigid to rubber-like), fractional free volume (a measure of how loosely molecules pack), thermal conductivity (the ability to transfer heat), density, and radius of gyration (a measure of molecular size). Ground-truth values are obtained from molecular dynamics simulations. 

\paragraph{Initial Algorithm} 
The graph rationalization method~\citep{liu2022graph} identifies subgraph structures, called “graph rationales,” and uses them for Graph Neural Network (GNN) predictions. To identify these rationales under limited supervision, \citet{liu2022graph} developed environment replacement, an augmentation that creates virtual examples in the latent space. It replaces the complementary structures of rationales (called environments) with others from the same training batch. Improving this method could strengthen both the generalizability and interpretability of GNNs for molecular property prediction. We use LLMs to read the paper~\citep{liu2022graph} and convert it to the input format we need as described in~\cref{subsec:method-init}.

\subsection{\usp}

\paragraph{Problem Description} 
The task is to measure semantic similarity between pairs of phrases drawn from patent documents~\citep{cenkci2022uspppm}. This is important for patent search and examination, where phrases with different wording (for example, “television set” and “TV set”) may have the same meaning, and where contextual knowledge (for example, what counts as a “strong material” in a given technical domain) is required. Each phrase pair is annotated with a similarity score between 0 (unrelated) and 1 (identical in meaning), and the technical domain is provided through the Cooperative Patent Classification system. Models are evaluated by the Pearson correlation between predicted and true similarity scores.

\paragraph{Initial Algorithm} 
The approach fine-tunes a BERT language model that has been pre-trained on patent text (``anferico/bert-for-patents'') with a regression layer added to predict similarity scores. Each training example is formed by concatenating the anchor phrase, the target phrase, and the technical context, separated by special tokens. The model is trained briefly and then evaluated by comparing predicted scores with the true similarity values using the Pearson correlation coefficient.

\section{Details on Experiment Results}

\subsection{Set-ups}

The user queries and hyperparameters in \method are shown in the list:

\begin{itemize}
    \item \cp \\
    User Query: \textit{You are an expert mathematician. Your task is to improve an algorithm that maximizes the sum of circle radii in the circle-packing problem within a unit square, using between 26 and 32 circles. Do not develop neural-network-based models. The algorithm must produce exact, valid packings that satisfy these constraints: circles do not overlap and remain entirely within the square.} \\
    Max iterations: 50
    
    \item \mt \\
    User Query: \textit{Your task is to significantly improve the model performance for converting molecular images to their InChI strings in the competition. You have a time budget of thirty minutes and access to an A6K GPU. The original method is intended for beginners, so make full use of available resources to improve it substantially as an expert in machine learning and chemistry. You can use pretrained models from transformers or from timm. Avoid placeholders for your method. Avoid warnings from Huggingface. For fair evaluation, avoid changing the deepevolve\_interface, run\_main\_with\_timeout, and get\_score functions. You can debug, but not subsample the test set to cheat the test performance.} \\
    Max iterations: 100

    \item \mol \\
    User Query: \textit{Your task is to improve the graph rationalization method for more accurate and interpretable molecular property prediction.} \\
    Max iterations: 100

    \item \nuclei \\
    User Query: \textit{Your task is to improve the nucleus detection models in a Kaggle competition within a compute budget of an A6k GPU with a maximum runtime of 30 minutes. You should significantly improve both the performance of the initial idea and its efficiency.} \\
    Max iterations: 50

    \item \ov \\
    User Query: \textit{Your task is to improve the nucleus detection models in a Kaggle competition within a compute budget of an A6k GPU with a maximum runtime of 30 minutes. You should gradually improve both the performance of the initial idea and its efficiency. For fair comparison: Do NOT change any code about the final evaluation such as the pred\_cols variable; You MUST use MCRMSELoss as the test\_criterion. You can define new criteria for training only. You can consider implementing the get\_bpps\_features() function to incorporate additional features. If you choose to use features beyond bpps, you may employ Hugging Face, but ensure those features are correctly added and not padded with placeholders or zeros.} \\
    Max iterations: 100 \\
    exploitation ratio: 0.8 \\
    elite selection ratio: 0.4 \\
    population size: 15 \\
    archive size: 5 \\
    number of islands: 3 \\
    migration interval:30 \\
    migration rate: 0.2

    \item \pd \\
    User Query: \textit{Your task is to improve the performance of the winning solution for the Kaggle competition on Parkinson disease progression prediction. You may propose a completely new approach that differs from the winning solution if you believe it will perform better.} \\
    Max iterations: 50

    \item \be \\
    User Query: \textit{Your task is to improve the solver for the partial differential equation (PDE). The solver should be applied to the Burgers equation with viscosity coefficients nu=1.0. Your computing budget is a 2080 Ti GPU with a maximum runtime of thirty minutes. Do not change the evaluation functions; Implement the `solver` function to solve the PDE. You must not modify the function signature. Please significantly reduce normalized root mean squared error (nRMSE), as well as achieve  higher convergence rate, and less computational time.} \\
    Max iterations: 200

    \item \pp \\
    User Query: \textit{Your task is to significantly improve polymer property prediction for five properties in the competition. The input SMILES strings are the monomer structures of polymers, using asterisks (*) to mark the polymerization points. Improve the initial idea by better incorporating polymerization inductive bias to reduce weighted MAE and increase $R^2$ for each property. Explore different ways to use polymer structures or properties and find the best. Your time budget is 30 minutes. Implement the idea within the time limit rather than creating a placeholder.} \\
    Max iterations: 50

    \item \usp \\
    User Query: \textit{Your task is to fine-tune Patent BERT to predict semantic similarity between phrase pairs from U.S. patents. Improve model performance, optimize training time and inference latency, and ensure the fixed three-epoch run finishes in thirty minutes. Focus solely on technical model and algorithm development. No legal-style assistance.} \\
    Max iterations: 50
\end{itemize}

\subsection{Summary of Algorithmic Evolution History}\label{sec:add-evolution-history}
\newcommand{\note}[1]{\textit{[Note: #1]}}

\paragraph{\mol} The algorithm progresses through auxiliary (contrastive, reconstruction) losses, motif-based, and adversarial learning strategies.
Version 1 establishes the foundation with contrastive learning on augmented rationale views, stabilized by adaptive loss reweighting. Version 2 enhances structural focus through motif-aware attribute masking, directing attention to chemically meaningful substructures. Version 3 further refines this by incorporating uncertainty-based soft motif selection, enabling the model to prioritize informative subgraphs dynamically. Version 4 strengthens representation fidelity with a self-supervised reconstruction objective that encourages the model to recover masked motifs. Version 5 introduces a dual-phase adversarial training schedule to improve model robustness and generalization under distribution shifts.

\paragraph{\mt} Version 1 uses a frozen ViT encoder and GPT-2 small decoder with molecule-aware tokenization to handle structured generation. Version 2 adds data augmentation such as rotation, shifting, and lighting perturbations for model training and grammar-constrained decoding. Version 3 and 4 train model with a dual loss combining cross-entropy and soft edit distance \note{The soft edit distance is a placeholder function in the code}. Version five implements a dynamic lambda scheduler to balance the competing loss objectives.

\paragraph{\cp} This algorithm evolves from basic geometric placement toward generating precise and guaranteed-valid solutions. Version 1 uses a structure called a power diagram to place circles without overlap, then refines their positions using optimization. Version 2 adds multiple starting points and more stable optimization techniques to improve reliability. Version 3 introduces small controlled adjustments to fix poor initial guesses and ensures that each circle stays within bounds. Version 4 improves how the method identifies neighboring circles and adds mathematical checks to certify that the final result fully satisfies the packing constraints.

\paragraph{\be}
The first stage (Versions 1–2) introduces an explicit Euler finite-difference solver with GPU acceleration and adaptive time stepping, later improved with error-based control and dense output for accuracy and snapshot recording.  
The second stage (Versions 3–4) transitions to a spectral method with IMEX-Euler time integration \note{Written in the code but not executed in the workflow}, integrating GPU kernel fusion and auto-tuned FFTs \note{Implemented as a placeholder function in the code} for faster and more accurate solutions.  
The third stage (Versions 5–7) focuses on advanced $\phi$-function evaluation (hybrid and rational Krylov), high-order Hermite interpolation, and refined adaptive stepping, forming a robust, high-precision spectral solver for the Burgers’ equation. \note{Written in the code but not executed in the workflow}

\paragraph{\pd} Versions 1–2 develop a Neural CDE model for continuous-time disease trajectory modeling. Versions 3-5 propose adaptive wavelet preprocessing for the time series data \note{Not implemented in the code}. Versions 6-7 incorporate meta-learning for rapid per-patient adaptation. Version 8 proposes a PINN-inspired regularization for biological consistency, and adaptive loss weighting to improve multi-objective training stability.

\paragraph{\nuclei} PointRend is introduced in version 1 to refine ambiguous segmentation boundaries. Versions 2, 3, and 5 introduce a calibrated uncertainty estimation module that refines only low-confidence regions to balance accuracy and computation. Version 3 enables early-exit to skip refinement for confident regions, with INT8 quantization applied for efficiency. Version 4 introduces self-distillation \note{Version 4 idea is not used because there is no teacher model}.

\paragraph{\ov} Versions 1–2 preprocess additional statistical features derived from RNA structure. Versions 3-6 add dynamic loss weighting to balance multiple degradation targets. Version 7 integrates self-supervised transformer embeddings into the node representations to enrich structural encoding \note{It is a placeholder function in the code}.

\paragraph{\pp} Versions 1–2 use dual-stage message passing to distinguish standard chemical bonds from polymer-specific periodic connections. A physics-informed auxiliary loss is added based on the degree of polymerization for glass transition temperature (Tg) prediction \note{However, the data is limited to one repeating unit only}. Versions 4-6 propose new ideas about BigSMILES parsing and property-specific pooling \note{ BigSMILES not supported, pooling not implemented in the code}. Versions 3 and 5 propose new ideas about meta-learning-based pooling (\note{Implemented but not used in the workflow}).

\paragraph{\usp} Versions 1–2 fine-tune Patent BERT using parameter-efficient LoRA with an ordinal regression head trained using smoothed BCE with logits and calibration for five ordinal similarity classes (0, 0.25, 0.5, 0.75, 1). Versions 3–4 introduce learnable CPC embeddings, fused into the latent space, and regularize the model using contrastive learning. Version 5 combines ordinal and contrastive losses in a dual-objective framework.

\subsection{\method Proposed Algorithm Code for the \mol Task}\label{sec:add-proposed-mol-code}

In \cref{fig:code_diff_mol_forward}, the new model forward function contains two additional components: the InfoNCE loss and the motif masking function. We present the complete code for these components in this subsection. Below is the code for the InfoNCE function:

\begin{minted}{diff}
+### >>> DEEPEVOLVE-BLOCK-START: Add InfoNCE loss for contrastive learning and ensure it is available in model.py
+### >>> DEEPEVOLVE-BLOCK-START: Update documentation for InfoNCE loss with advanced negative sampling note
+### >>> DEEPEVOLVE-BLOCK-START: Update InfoNCE loss to support uncertainty-guided negative sampling
+def info_nce_loss(z1, z2, temperature=0.5, negatives=None):
+    """
+    Computes the InfoNCE loss using current batch negatives.
+    If 'negatives' is provided, applies advanced negative sampling for enhanced robustness.
+    """
+    z1 = torch.nn.functional.normalize(z1, p=2, dim=1)
+    z2 = torch.nn.functional.normalize(z2, p=2, dim=1)
+    if negatives is not None:
+        negatives = torch.nn.functional.normalize(negatives, p=2, dim=1)
+        sim_pos = torch.sum(z1 * z2, dim=1, keepdim=True) / temperature
+        sim_neg = torch.matmul(z1, negatives.t()) / temperature
+        logits = torch.cat([sim_pos, sim_neg], dim=1)
+        labels = torch.zeros(z1.size(0), device=z1.device, dtype=torch.long)
+        loss = torch.nn.functional.cross_entropy(logits, labels)
+    else:
+        logits = torch.matmul(z1, z2.t()) / temperature
+        labels = torch.arange(z1.size(0), device=z1.device)
+        loss = torch.nn.functional.cross_entropy(logits, labels)
+    return loss
+### <<< DEEPEVOLVE-BLOCK-END
+### <<< DEEPEVOLVE-BLOCK-END
+### <<< DEEPEVOLVE-BLOCK-END
\end{minted}

Here is the code for the motif masking function:

\begin{minted}{diff}
+    ### >>> DEEPEVOLVE-BLOCK-START: Add motif-aware attribute masking method to GraphEnvAug
+    ### >>> DEEPEVOLVE-BLOCK-START: Update motif_mask for uncertainty-aware differentiable motif extraction using Gumbel-Softmax and MC Dropout
+    def motif_mask(self, batched_data):
+        import copy
+        import torch.nn.functional as F
+
+        # motif_mask: compute adaptive motif mask without altering original x
+        new_data = copy.deepcopy(batched_data)
+        orig_x = new_data.x
+        x_float = orig_x.float()
+
+        # Initialize motif_selector and dropout if not already defined
+        if not hasattr(self, "motif_selector"):
+            self.motif_selector = torch.nn.Linear(orig_x.size(1), 2).to(orig_x.device)
+            self.motif_dropout = torch.nn.Dropout(p=0.5)
+        num_samples = (
+            self.mc_dropout_samples if hasattr(self, "mc_dropout_samples") else 5
+        )  # Use configured number of MC dropout samples
+        motif_samples = []
+        tau = 1.0  # Temperature parameter for Gumbel-Softmax; can be tuned
+        for _ in range(num_samples):
+            logits = self.motif_selector(x_float)
+            logits = self.motif_dropout(logits)  # MC Dropout
+            sample = F.gumbel_softmax(logits, tau=tau, hard=False, dim=1)[
+                :, 1
+            ].unsqueeze(1)
+            motif_samples.append(sample)
+        motif_samples = torch.stack(
+            motif_samples, dim=0
+        )  # Shape: [num_samples, num_nodes, 1]
+        mean_score = motif_samples.mean(dim=0)  # Aggregated motif probability
+        uncertainty = motif_samples.var(dim=0)  # Variance as uncertainty
+        threshold_uncertainty = 0.05  # Adaptive threshold hyperparameter
+        adaptive_mask = torch.where(
+            uncertainty < threshold_uncertainty,
+            mean_score,
+            mean_score * (threshold_uncertainty / (uncertainty + 1e-8)),
+        )
+
+        # Store computed uncertainty for potential adversarial perturbation
+        self.last_uncertainty = uncertainty
+        # DEBUG: store adaptive mask for use in GNN (applied in conv.py)
+        new_data.mask = adaptive_mask
+        return new_data
+### <<< DEEPEVOLVE-BLOCK-END
+### <<< DEEPEVOLVE-BLOCK-END
\end{minted}

\end{document}